\documentclass[nai]{iosart2x}

\usepackage{graphicx,subcaption}
\usepackage{amsmath}
\usepackage{amsfonts}
\usepackage{booktabs}
\usepackage{todonotes}
\usepackage{semantic}
\usepackage{multirow}
\usepackage{array}
\usepackage[section]{placeins}
\usepackage{natbib}
\usepackage{algpseudocode}
\usepackage{algorithm}
\usepackage{hyperref}
\usepackage{caption}
\usepackage{subcaption}
\usepackage{cleveref}
\usepackage{hhline}
\usepackage{adjustbox}

\newcommand{\el}{$\mathcal{EL}^{++}$}
\newcommand{\interp}[1]{{#1}^\mathcal{I}}
\newcolumntype{C}[1]{>{\centering\let\newline\\\arraybackslash\hspace{0pt}}m{#1}}

\pubyear{0000}
\volume{0}
\firstpage{1}
\lastpage{1}

\begin{document}

\numberlinesfalse

\begin{frontmatter}

\title{DELE: Deductive \el \thinspace Embeddings for Knowledge Base Completion}
\runtitle{DELE}
\subtitle{NeSy 2024 Special Issue, paper \# 51}


\begin{aug}
\author[A]{\inits{O.}\fnms{Olga}
  \snm{Mashkova}\ead[label=e1]{olga.mashkova@kaust.edu.sa}
}
\author[A]{\inits{F.}\fnms{Fernando} \snm{Zhapa-Camacho}\ead[label=e2]{fernando.zhapacamacho@kaust.edu.sa}}
\author[A,B,C]{\inits{R.}\fnms{Robert} \snm{Hoehndorf}\ead[label=e3]{robert.hoehndorf@kaust.edu.sa}\thanks{Corresponding author. \printead{e3}.}}
\address[A]{Computer Science Program, Computer, Electrical, and
  Mathematical Sciences \& Engineering Division, \orgname{King Abdullah
  University of Science and Technology},
Thuwal, \cny{Saudi Arabia}}
\address[B]{KAUST Center of Excellence for Smart Health, \orgname{King Abdullah
  University of Science and Technology},
Thuwal, \cny{Saudi Arabia}}
\address[C]{KAUST Center of Excellence for Generative AI, \orgname{King Abdullah
  University of Science and Technology},
Thuwal, \cny{Saudi Arabia}\printead[presep={\\}]{e3}}

\end{aug}


\begin{abstract}

Ontology embeddings map classes, roles, and individuals in
ontologies into $\mathbb{R}^n$, and within $\mathbb{R}^n$ similarity
between entities can be computed or new axioms inferred.
For ontologies in the Description Logic $\mathcal{EL}^{++}$, several
optimization-based embedding methods have been developed
that explicitly generate models of an ontology.
However, these methods suffer from some limitations;
they do not distinguish between statements that are unprovable and
provably false, and therefore they may use entailed statements as
negatives. Furthermore, they do not utilize the deductive closure of
an ontology to identify statements that are inferred but not asserted.
We evaluated a set of embedding methods for $\mathcal{EL}^{++}$
ontologies, incorporating several modifications that aim to make use
of the ontology deductive closure.  In particular, we designed novel
negative losses that account both for the deductive closure and
different types of negatives and formulated evaluation methods for
knowledge base completion.  We demonstrate that our embedding methods
improve over the baseline ontology embedding in the task of knowledge
base or ontology completion.
\end{abstract}


\end{frontmatter}

\section{Introduction}

Several methods have been developed to embed Description Logic
theories or ontologies in vector
spaces~\cite{owl2vecstar,dl2vec,kulmanov2019embeddings,mondal2021emel++,peng2022description,xiong2022faithful,jackermeier2023box,ozcep2023embedding}. These
embedding methods preserve some aspects of the semantics in the vector
space, and may enable the computation of semantic similarity,
inferring axioms that are entailed, and predicting axioms that are not
entailed but may be added to the theory. For the lightweight
Description Logic \el, several geometric embedding methods have been
developed~\cite{kulmanov2019embeddings,mondal2021emel++,xiong2022faithful,jackermeier2023box,ozcep2023embedding}. They
can be proven to ``faithfully'' approximate a model in the sense that,
if a certain optimization objective is reached (usually, a loss
function reduced to $0$), the embedding method has constructed a model
of the \el theory. Geometric model construction enables the execution
of various tasks. These tasks include knowledge base completion and
subsumption prediction via either testing the truth of a statement
under consideration in a single (approximate) model or aggregating
truth values over multiple models.

Advances on different geometric embedding methods have usually focused
on the expressiveness of the embedding methods; originally,
hyperballs~\cite{kulmanov2019embeddings} where used to represent the
interpretation of concept symbols, yet hyperballs are not closed under
intersection. Therefore, axis-aligned boxes were
introduced~\cite{peng2022description,xiong2022faithful,jackermeier2023box}. Furthermore,
\el allows for axioms pertaining to roles, and several methods
have extended the way in which roles are modeled~\cite{jackermeier2023box,kulmanov2019embeddings,xiong2022faithful}. However,
there are several aspects of geometric embeddings that have not yet
been investigated. In particular, for \el, there are sound and
complete reasoners with efficient implementations that scale to very
large knowledge bases~\cite{elk}; it may therefore be possible to
utilize a deductive reasoner together with the embedding process to
improve generation of embeddings that represent geometric models.

We evaluate geometric embedding methods and incorporate deductive
inference into the training process.  We use the {\it
  ELEmbeddings}~\cite{kulmanov2019embeddings}, {\it
  ELBE}~\cite{peng2022description}, and
$Box^2EL$~\cite{jackermeier2023box} models for our experiments;
however, our results also apply to other geometric embedding methods
for \el.

Our main contributions are as follows:
\begin{itemize}
\item We propose loss functions that incorporate negative samples in
  all normal forms and account for deductive closure during training.
\item We introduce a fast approximate algorithm for computing the
  deductive closure of an \el theory and use it to improve negative
  sampling during model training.
\item We formulate evaluation methods for knowledge base completion
  that account for the deductive closure during evaluation.
\end{itemize}

This is an extended version of our previous
work~\cite{mashkova2024enhancing}.  Here, we include a more
comprehensive treatment of computing the deductive closure and using
the deductive closure with \el embedding methods (e.g., for model evaluation). We include
  additional experiments on three benchmark datasets: besides protein
  function prediction and protein--protein interaction, we study the
  subsumption prediction task on Food
  Ontology~\cite{dooley2018foodon} and GALEN ontology~\cite{rector1996galen}. Furthermore, we consider two
  additional geometric ontology embedding models, {\it
    ELBE}~\cite{peng2022description} and
  $Box^2EL$~\cite{jackermeier2023box}, apart from {\it
    ELEmbeddings}~\cite{kulmanov2019embeddings} for which we also extend our methods. We make our code and
data available at
\url{https://github.com/bio-ontology-research-group/DELE}.

\section{Preliminaries}

\subsection{Description Logic \el} \label{dl}

Let $\Sigma = (\mathbf{C}, \mathbf{R}, \mathbf{I})$ be a signature
with set $\mathbf{C}$ of concept names, $\mathbf{R}$ of role names,
and $\mathbf{I}$ of individual names. Given $r \in \mathbf{R}$, and $a, b \in \mathbf{I}$, \el concept
descriptions are constructed with the grammar $ \bot \mid \top \mid C \sqcap D \mid \exists r.C \mid \{a\}$ where $C, D$ are \el concept descriptions and $r$ is a role name. ABox axioms are of the form $C(a)$
and $r(a,b)$, TBox axioms are of the form $C \sqsubseteq D$, and RBox axioms are of the form
$r_1 \circ r_2 \circ \dots \circ r_n \sqsubseteq r$. \el \emph{generalized concept inclusions} (GCIs) and \emph{role inclusions} (RIs) can be normalized to follow one of the forms listed in Table~\ref{acronym_table}~\cite{baader2005pushing}. If an ontology contains a non-empty ABox, then, first, concept assertion axioms of the form $C(a)$ are processed: for each complex concept description $C$, a new concept name $A$ is introduced and axiom $C \equiv A$ is added to the TBox~\cite{penaloza2008instance}. Normalization rules for TBox axioms~\cite{baader2005pushing} can be found in Table~\ref{norm_rules_table}. They preserve the same models for ontologies~\cite{mendez2011classification} and the normalized ontology is a conservative extension of the original
  ontology~\cite{suntisrivaraporn2009polynomial}. Following previous
  works~\cite{kulmanov2019embeddings,peng2022description,jackermeier2023box}
  we develop our methodology specifically for normalized \el
  ontologies. The advantage of normalizing \el ontologies is that
  their deductive closure is finite whereas the deductive closure of
  non-normalized \el ontologies is not. Therefore, in our work, when
  we refer to the language \el, we always intend the normalized \el,
  i.e., where all valid axiom types are those listed in
  Table~\ref{acronym_table}.
  
\begin{table}[!h] \centering
\caption{Normalized forms of \el generalized concept
    inclusions (GCIs) and role inclusions
    (RIs). Here, $r, r_1, r_2, s$ are role names, $A, B, E$ are concept names.} \label{acronym_table}
\begin{tabular}{| C{2cm} | C{2cm} |}
\hline
{\bf Acronym} & {\bf Axiom type} \\
\hline
GCI0 & $A \sqsubseteq B$ \\
\hline
GCI1 & $A \sqcap B \sqsubseteq E$ \\
\hline
GCI2 & $A \sqsubseteq \exists r.B$ \\
\hline
GCI3 & $\exists r.A \sqsubseteq B$ \\
\hline
GCI0-BOT & $A \sqsubseteq \bot$ \\
\hline
GCI1-BOT & $A \sqcap B \sqsubseteq \bot$ \\
\hline
GCI3-BOT & $\exists r.A \sqsubseteq \bot$ \\
\hline
RI0 & $r \sqsubseteq s$ \\
\hline
RI1 & $r_1 \circ r_2 \sqsubseteq s$ \\
\hline
\end{tabular}
\end{table}

\begin{table}
\caption{Normalization rules for TBox for \el based on Baader et al.~\cite{baader2005pushing}. Here, $r, r_1, \dots, r_k, s$ are role names, $C, D, B$ are arbitrary concept descriptions,
    $\hat{C}, \hat{D}$ are complex concept descriptions (i.e., not
    concept names), $A$ is a fresh concept name
    and $u$ is a fresh role name.} \label{norm_rules_table}
\centering
\begin{tabular}{| C{5cm} | C{5cm} |}
\hline
{\bf Input} & {\bf Output} \\
\hline
$r_1 \circ \cdots \circ r_k \sqsubseteq s$ & $r_1 \circ \cdots \circ r_{k-1} \sqsubseteq u, \thinspace u \circ r_k \sqsubseteq s$ \\
\hline
$C \sqcap \hat{D} \sqsubseteq E$ & $\hat{D} \sqsubseteq A, \thinspace C \sqcap A \sqsubseteq E$ \\
\hline
$\exists r.\hat{C} \sqsubseteq D$ & $\hat{C} \sqsubseteq A, \thinspace \exists r.A \sqsubseteq D$ \\
\hline
$\bot \sqsubseteq D$ & $\emptyset$ \\
\hline
$\hat{C} \sqsubseteq \hat{D}$ & $\hat{C} \sqsubseteq A, \thinspace A \sqsubseteq \hat{D}$ \\
\hline
$B \sqsubseteq \exists r.\hat{C}$ & $B \sqsubseteq \exists r.A, \thinspace A \sqsubseteq \hat{C}$ \\
\hline
$B \sqsubseteq C \sqcap D$ & $B \sqsubseteq C, \thinspace B \sqsubseteq D$ \\
\hline
\end{tabular}
\end{table}

To define the semantics of an $\mathcal{EL}^{++}$ theory, we
use~\cite{baader2005pushing} an \emph{interpretation domain}
$\Delta^{\mathcal{I}}$ and an \emph{interpretation function}
$\cdot^{\mathcal{I}}$.  For every concept $A \in \mathbf{C}$ $\interp{A} \subseteq \interp{\Delta}$; 
individual $a \in \mathbf{I}$,
$\interp{a} \in \interp{\Delta}$; role $r \in \mathbf{R}$,
$\interp{r} \subseteq \interp{\Delta} \times \interp{\Delta}$. Furthermore,
the semantics for other $\mathcal{EL}^{++}$ constructs are the
following (omitting concrete domains and role inclusions):
\begin{equation}
  \nonumber
  \begin{split}
    {\bot}^{\mathcal{I}} &= \emptyset\\
    {\top}^{\mathcal{I}} &=\Delta^{\mathcal{I}},\\
    (C \sqcap D)^{\mathcal{I}} &= C^{\mathcal{I}} \cap D^{\mathcal{I}}, \\
    (\exists r . C)^{\mathcal{I}} &=\left\{x \in \Delta^{\mathcal{I}}
    \mid \exists \thinspace y \in \Delta^{\mathcal{I}} : ((x, y) \in r^{\mathcal{I}} \land y \in \interp{C})\right\},\\
   {\{a\}}^{\mathcal{I}} &= \{a^{\mathcal{I}}\}
  \end{split}
  \quad \quad
  \begin{split}
   \end{split}
\end{equation}

An interpretation $\mathcal{I}$ is a model for an axiom ${C} \sqsubseteq {D}$ if and only if $\interp{C} \subseteq \interp{D}$, for an axiom $C(a)$ if and only if $\interp{a} \in \interp{C}$; and for an axiom $r(a,b)$ if and only if $(\interp{a},\interp{b}) \in \interp{r}$~\cite{dl_handbook}. The relation of semantic entailment, $\models$, is defined as a relation between a theory $\mathcal{T}$ and axiom $\phi$: $\mathcal{T} \models \phi$ iff every model of $\mathcal{T}$ is also a model of $\phi$ ($Mod(\mathcal{T}) \subseteq Mod(\{\phi\})$)~\cite{tarski1936concept}.

\subsection{Knowledge Base Completion}

The task of knowledge base completion is the addition (or prediction)
  of axioms which hold yet are not represented in the knowledge base.
We call the task ``ontology completion'' when exclusively TBox axioms
are predicted. The task of knowledge base completion may encompass
both deductive~\cite{sato2018deductive,jiang2012combining} and
inductive~\cite{bouraoui2017inductive,d2012ontology} inference
processes and give rise to two subtly different tasks: adding only
``novel'' axioms to a knowledge base that are {\em not} in the
deductive closure of the knowledge base, and adding axioms that are in
the deductive closure as well as some ``novel'' axioms that are not
deductively inferred; both tasks are related but differ in how they
are evaluated.

Inductive inference, analogously to knowledge graph
completion~\cite{chen2020knowledge}, predicts axioms based on patterns
and regularities within the knowledge base. Knowledge base completion,
or ontology completion, can be further distinguished based on the
information that is used to predict ``novel'' axioms.  We distinguish
between two approaches to knowledge base completion: (1) knowledge
base completion which relies solely on (formalized) information within
the knowledge base to predict new axioms, and (2) knowledge base
completion which incorporates side information, such as text, to
enhance the prediction of new axioms. Here, we mainly consider the
first case.

  \subsection{Deductive Closure}
  The {\it deductive closure} of a theory $T$ refers to the smallest
  set containing all statements which can be inferred by deductive
  reasoning over $T$; for a given deductive relation $\vdash$, we call
  $T^\vdash = \{ \phi \thinspace | \thinspace T \vdash \phi \}$ the
  deductive closure of $T$.  In knowledge bases, the deductive closure
  is usually not identical to the asserted axioms in the knowledge
  base; it is also usually infinite. Representing the deductive
  closure is challenging since it is infinite, but, in \el, any
  knowledge base can be normalized to one of the seven normal forms;
  therefore, we can compute the deductive closure with respect to
  these normal forms, and this set will be finite (as long as the sets
  of concept and role names are finite). For example, all entailed axioms of type $A \sqsubseteq B$ will be a subset of the set of all possible axioms of GCI0 type having cardinality $|{\bf C}|^2$ where $|{\bf C}|$ is the cardinality of the set of all concept names. Similarly, the cardinality of GCI0-BOT deductive closure will be limited by $|{\bf C}|$, GCI1 deductive closure cardinality will be limited by $|{\bf C}|^3$, and GCI1-BOT
    deductive closure cardinality by $|{\bf C}|^2$ since one of the
    concepts is fixed. GCI2 and GCI3 deductive closures cardinality
    will depend on the total number of 
    roles $|{\bf R}|$ and will
    be limited by $|{\bf C}|^2 \cdot |{\bf R}|$, and, finally, the
    number of entailed axioms of GCI3-BOT type will not exceed
    $|{\bf C}| \cdot |{\bf R}|$.

\section{Related Work}

\subsection{Graph-Based Ontology Embeddings}
Graph-based ontology embeddings rely on a construction (projection) of
graphs from ontology axioms mapping ontology classes, individuals and
roles to nodes and labeled edges~\cite{graph_projections}. Embeddings
for nodes and edge labels are optimized following two strategies: by
generating random walks and using a sequence learning method such as
Word2Vec~\cite{word2vec} or by using Knowledge Graph Embedding (KGE)
methods~\cite{wang2017knowledge}.  These type of methods have been
shown effective on knowledge base and ontology
completion~\cite{owl2vecstar} and have been applied to domain-specific
tasks such as protein--protein interaction
prediction~\cite{owl2vecstar} or gene--disease association
prediction~\cite{dl2vec,embedpvp}.  Graph-based methods rely on
adjacency information of the ontology structure but cannot easily
handle logical operators and do not approximate ontology models.
Therefore, graph-based methods are not ``faithful'', i.e., do not
approximate models, do not allow determining whether statements are
``true'' in these models, and therefore cannot be used to perform
semantic entailment.

\subsection{Geometric-Based Ontology Embeddings}

Multiple methods have been developed for the geometric construction of
models for the $\mathcal{EL}^{++}$ language.
ELEmbeddings~\cite{kulmanov2019embeddings} constructs an
interpretation of concept names as sets of points lying within an open
$n$-dimensional ball and generates an interpretation of role names as
the set of pairs of points that are separated by a vector in
$\mathbb{R}^n$, i.e., by the embedding of the role name.
EmEL++~\cite{mondal2021emel++} extends ELEmbeddings with more
expressive constructs such as role chains and role
inclusions. Another extension of ELEmbeddings,
  EmELvar~\cite{mohapatra2021emelvar}, introduces role embeddings
  which enables handling many-to-many 
  roles and provides a
  perspective to extending the method to more expressive Description
  Logics. ELBE~\cite{peng2022description} and
BoxEL~\cite{xiong2022faithful} use $n$-dimensional axis-aligned boxes
to represent concepts, which has an advantage over balls because the
intersection of two axis-aligned boxes is a box whereas the
intersection of two $n$-balls is not an $n$-ball. BoxEL additionally
preserves ABox facilitating a more accurate representation of
knowledge base's logical structure by ensuring, e.g., that an entity
has the minimal volume.  Box$^2$EL~\cite{jackermeier2023box}
represents ontology roles more expressively with two boxes encoding
the semantics of the domain and codomain of roles. Box$^2$EL enables
the expression of one-to-many roles as opposed to other methods. The box-based method
  TransBox~\cite{yang2025transbox} aims to effectively capture all
  $\mathcal{EL}^{++}$ logical operations allowing for precise
  representation of any arbitrarily complex concept description. The
  fact that any normalized 
  $\mathcal{EL}^{++}$ theory 
  has a finite deductive closure allows the definition of a canonical
  model where all and only entailed axioms are true. Lacerda et al.~\cite{lacerda2024faithel,lacerda2024strong} developed FaithEL, a strongly TBox-faithful method based on convex sets in $n$-dimensional real-valued space which interprets concepts and roles as subsets of unitary hypercubes. FaithEL constructs a ``canonical'' model of an ontology and is able to predict new assertions consistent with a TBox. Axis-aligned cone-shaped geometric model introduced in~\cite{ozcep2023embedding}
deals with $\mathcal{ALC}$ ontologies and allows for full negation of
concepts and existential quantification by construction of convex sets
in ${\mathbb R}^n$. This work has not yet been implemented or
evaluated in an application.

\subsection{Knowledge Base Completion Task}

Several recent advancements in the knowledge base completion rely on
side information as included in Large Language Models (LLMs).
Ji et al.~\cite{ji2023ontology} explores how pretrained language models can be
utilized for incorporating one ontology into another, with the main
focus on inconsistency handling and ontology
coherence. HalTon~\cite{cao2023event} addresses the task of event
ontology completion via simultaneous event clustering, hierarchy
expansion and type naming utilizing BERT~\cite{devlin2018bert} for
instance encoding. Li et al.~\cite{li2024ontology} formulates knowledge base
completion task as a Natural Language Inference (NLI) problem and
examines how this approach may be combined with concept embeddings for
identifying missing knowledge in ontologies. As for other approaches,
Mežnar et al.~\cite{mevznar2022ontology} proposes a method that converts an ontology
into a graph to recommend missing edges using structure-only link
analysis methods, Shiraishi et al.~\cite{shiraishi2024self} constructs matrix-based
ontology embeddings which capture the global and local information for
subsumption prediction. All these methods use side information from
LLMs and would not be applicable, for example, in the case where a
knowledge base is private or consists of only identifiers; we do not
consider methods based on pre-trained LLMs here as baselines.

\section{Negative sampling and Objective Functions} \label{losses}

Currently available geometric ontology embedding models
  which construct a model of an ontology by optimizing some objective
  function usually sample negative examples during training
phase~\cite{kulmanov2019embeddings,peng2022description,jackermeier2023box,mondal2021emel++,mohapatra2021emelvar,yang2025transbox}. This
  operation prevents overgeneralization of learned embeddings and
  trivial satisfiability in case a model
  collapses~\cite{yang2025transbox,kulmanov2019embeddings} by
  incorporating additional constraints within a model. Ontology
embedding methods select negatives by replacing one of the concepts
with a randomly chosen one (either from the set of all concept names,
or a subset thereof). 
{\it ELEmbeddings}, {\it ELBE} and $Box^2EL$ use a single loss for
``negatives'', i.e., axioms that are not included in the knowledge
base; the loss is used only for axioms of the form
$A \sqsubseteq \exists r.B$ (GCI2) which are randomly sampled;
negatives are not sampled for other normal forms.  Correspondingly,
the embedding methods were primarily evaluated on predicting GCI2
axioms ($Box^2EL$ was also evaluated on subsumption prediction); this
evaluation procedure might have introduced biases towards axioms of
type GCI2, and influenced the ability of geometric models to predict
axioms of other types. Specifically, the lack of negative
  examples of other axiom types may lead to geometric models in which
  many axioms are true even if they are not entailed, leading to a
  decreased ability to find axioms that can be added to a theory in
  the task of knowledge base completion.  Consequently, we also
sample negatives for other normal forms and add ``negative'' losses
(i.e., losses for the sampled ``negatives'') for all other normal
forms.

\subsection{ELEmbeddings Negative Losses} \label{elem_losses}

For negative loss construction in {\it ELEmbeddings}, we
  employ notations from the {\it ELEmbeddings} method where
  $r_{\eta}(a), \thinspace r_{\eta}(b), \thinspace r_{\eta}(e)$ 
  and $f_{\eta}(a), \thinspace f_{\eta}(b), \thinspace f_{\eta}(e)$ 
  denote the radius and the ball center associated with classes 
  $a, b, e$, respectively; and $f_{\eta}(r_0)$ denotes the embedding vector associated with 
  role $r$. $\gamma$ stands for a margin
  parameter, and $\varepsilon$ is a small positive number. There is a
  geometric part as well as a regularization part for each new
  negative loss forcing class centers to lie on a unit
  $\ell_2-$sphere.

  For {\it ELEmbeddings}, as reflected in Eq.~\ref{eq1}, we use the
  original GCI1-BOT loss for disjoint classes; although
  non-containment of ball corresponding to $A$ within the ball corresponding to B is not equivalent to their disjointness, the
  loss aims to minimize the classes' overlap for better optimization:
  
  \begin{equation} \label{eq1}
    loss_{A \not\sqsubseteq B}(a, b) = \max(0, r_{\eta}(a) +
    r_{\eta}(b) - \|f_{\eta}(a) - f_{\eta}(b))\| + \gamma) + |
    \|f_{\eta}(a)\| - 1| + | \|f_{\eta}(b)\| - 1|
  \end{equation}
  
  The same logic applies for the negative loss in Eq.~\ref{eq3} where
  we minimize overlap between the translated ball corresponding to class $A$ and the ball representing $B$:

  \begin{align} \label{eq3}
    \begin{split}
      loss_{\exists r.A\not\sqsubseteq B}(r_0, a, b) = \max(0,
      r_{\eta}(a) + r_{\eta}(b) - \|f_{\eta}(a) - f_{\eta}(r_0) -
      f_{\eta}(b))\| + \gamma) + \\ + | \|f_{\eta}(a)\| - 1| + |
      \|f_{\eta}(b)\| - 1|
    \end{split}
  \end{align}

  Negative loss~\ref{eq2} is constructed similarly to the
  $A \sqcap B \sqsubseteq E$ loss: the first part penalizes
  non-overlap of the classes $A$ and $B$ (we do not consider the
  disjointness case since, for every class $X$, we have
  $\bot \sqsubseteq X$); furthermore, for negative sampling of axioms
  of this type, we vary only the $E$ part of GCI1 axioms from the
  ontology, so the intersection of $A$ and $B$ is non-empty by
  assumption:

  \begin{align} \label{eq2}
    \begin{split}
      loss_{A \sqcap B \not\sqsubseteq E}(a, b, e) = \max(0, -
      r_{\eta}(a) - r_{\eta}(b) + \|f_{\eta}(a) - f_{\eta}(b))\| -
      \gamma) + \\ + \max(0, r_{\eta}(a) - \|f_{\eta}(a) -
      f_{\eta}(e))\| + \gamma) + \max(0, r_{\eta}(b) - \|f_{\eta}(b) -
      f_{\eta}(e))\| + \gamma) + \\ + | \|f_{\eta}(a)\| - 1| + |
      \|f_{\eta}(b)\| - 1| + | \|f_{\eta}(e)\| - 1|
    \end{split}
  \end{align}

  The second and the third part force the center corresponding to $E$
  not to lie in the intersection of balls associated with $A$ and
  $B$. Here we do not consider constraints on the radius of the ball
  for the $E$ class and focus only on the relative positions of the
  $A, B$ and $E$ class centers and the overlapping of $n$-balls representing $A$ and $B$. Since the first part of the loss
  encourages classes to have a non-empty intersection, we use it as a
  negative loss for GCI1-BOT axioms:

  \begin{equation} \label{eq5} 
  loss_{A \sqcap B \not\sqsubseteq
      \bot}(a, b) = \max(0, - r_{\eta}(a) - r_{\eta}(b) +
    \|f_{\eta}(a) - f_{\eta}(b))\| - \gamma) + | \|f_{\eta}(a)\| - 1|
    + | \|f_{\eta}(b)\| - 1|
  \end{equation}

  In the original method losses for axioms of type GCI0-BOT and
  GCI3-BOT force radii of unsatisfiable classes to become $0$. For the
  correspondent negative losses (see Eq.~\ref{eq4} and Eq.~\ref{eq6})
  we use the interpretation for satisfiable classes as balls with
  non-zero radius, i.e., with a radius which equals to or greater than
  some small positive number $\varepsilon$:
  \begin{align}
    loss_{A \not\sqsubseteq \bot}(a) &= \max(0, \varepsilon - r_{\eta}(a))  \label{eq4} \\
    loss_{\exists r.A \not\sqsubseteq \bot}(r_0, a) &= \max(0, \varepsilon - r_{\eta}(a)) \label{eq6}
  \end{align}

\subsection{ELBE Negative Losses} \label{elbe_losses}

{\it ELBE} is a model that relies on boxes instead of balls.
  Here, similarly, $\varepsilon$ is a small positive number, $e_c(a), \thinspace e_c(b)$ and 
  $e_o(a), \thinspace e_o(b)$ denote the box center and the box offset associated with classes $a, b$, respectively, 
  $e_c(r_0)$ denotes the embedding vector associated with
  role $r$, and $e_c(new), \thinspace e_o(new)$ correspond to the
  center and the offset of the box which is the result of intersection
  of boxes associated with concepts $a$ and $b$, $margin$ stands for a margin parameter.

  Following the same method of negative loss construction for {\it ELEmbeddings}, we use GCI1-BOT loss as a negative loss for
  $A \sqsubseteq B$ axioms (see Eq.~\ref{elbe_eq1}):
  \begin{equation} \label{elbe_eq1} 
  loss_{A \not\sqsubseteq B}(a, b) =
    \|\max(zeros, -|e_c(a) - e_c(b)| + e_o(a) + e_o(b) + margin)\|
  \end{equation}

  Since axis-aligned hyperrectangles are closed under intersection, we
  also use GCI1-BOT for the intersection of boxes representing $A$ and
  $B$ concepts and the $E$ box (see Eq.~\ref{elbe_eq2}):
  \begin{equation} \label{elbe_eq2} 
  loss_{A \sqcap B \not\sqsubseteq
      E}(a, b, e) = \|\max(zeros, -|e_c(new) - e_c(e)| + e_o(new) +
    e_o(e) + margin)\|
  \end{equation}

  This property also allows us to interpret each negative sample for
  $A \sqcap B \sqsubseteq \bot$ axioms as a box intersection with
  nonzero offset (see Eq.~\ref{elbe_eq5}):
  \begin{equation} \label{elbe_eq5} 
  loss_{A \sqcap B \not\sqsubseteq
      \bot}(a, b) = \max(0, \varepsilon - \|e_o(new)\|)
  \end{equation}

  Other negative losses have the form similar to the ones constructed for {\it ELEmbeddings}:
  \begin{align} 
    loss_{\exists r.A \not\sqsubseteq B}(r_0, a, b) &= \|max(zeros, -|e_c(a) - e_c(r_0) - e_c(b))| + e_o(a) + e_o(b) + margin)\| \label{elbe_eq3} \\
    loss_{A \not\sqsubseteq \bot}(a) &= \max(0, \varepsilon - \|e_o(a)\|) \label{elbe_eq4} \\
    loss_{\exists r.A \not\sqsubseteq \bot}(r_0, a) &= \max(0, \varepsilon - \|e_o(a)\|) \label{elbe_eq6}
  \end{align}

\subsection{$Box^2EL$ Negative Losses} \label{box2el_losses}

$Box^2EL$ is also based on boxes but uses a different role modeling compared to ELBE. Additionally making use of the
  notations from $Box^2EL$~\cite{jackermeier2023box}, $\varepsilon$ is
  a small positive number, $Box(A), \thinspace Box(B)$, $Box(E)$ are boxes associated with classes 
  $a, b, e$, respectively, $\gamma$ denotes a margin parameter, $\delta$ is a parameter from the GCI2
  negative loss, $Head(r)$ represents the head box of role $r$ interpretation, and $Bump(A)$ 
  corresponds to a bump vector associated with concept $A$.

  Equations~\ref{box2el_eq1} and \ref{box2el_eq2} are constructed in a
  similar fashion as for {\it ELBE} based on the GCI1-BOT loss which
  penalizes the element-wise distance $\boldsymbol{d}$ between
  axis-aligned boxes:
  \begin{align}
    loss_{A \not\sqsubseteq B}(a, b) &= \|\max(\boldsymbol{0}, -(\boldsymbol{d}(Box(A), Box(B)) + \gamma))\| \label{box2el_eq1} \\
    loss_{A \sqcap B \not\sqsubseteq E}(a, b, e) &= \|\max(\boldsymbol{0}, -(\boldsymbol{d}(Box(A) \cap Box(B), Box(E)) + \gamma))\|    \label{box2el_eq2}
  \end{align}

  Negative losses~\ref{box2el_eq4}--\ref{box2el_eq6} encourage boxes
  to be non-empty:
  \begin{align}
    loss_{A \not\sqsubseteq \bot}(a) &= \max(0, \varepsilon - \|o(A)\|) \label{box2el_eq4} \\
    loss_{A \sqcap B \not\sqsubseteq \bot}(a, b) &= \max(0, \varepsilon - \|o(Box(A) \cap Box(B))\|) \label{box2el_eq5} \\
    loss_{\exists r.A \not\sqsubseteq \bot}(r_0, a) &= \max(0, \varepsilon - \|o(A)\|) \label{box2el_eq6}
  \end{align}

  The GCI3 negative loss reflects the structure of the original GCI3
  loss:
  \begin{equation} \label{box2el_eq3} 
  loss_{\exists r.A\not\sqsubseteq B}(r_0, a, b) = (\delta - \mu(Head(r) - Bump(A), Box(B)))^2
  \end{equation}

\begin{algorithm}[h]\captionsetup{labelfont={sc,bf}, labelsep=newline} 
  \caption{An algorithm for computation of axioms in the deductive
    closure using inference rules; axioms in bold correspond to
    subclass/superclass axioms derived using ELK reasoner (here we use
    the transitive closure of the ELK inferences); plain axioms come
    from the knowledge base. The input of the algorithm is
      the set of normalized axioms of the knowledge base, the output
      is the set of entailed axioms computed with respect to all
      normal forms according to syntactic rules stated in the
      algorithm.}
\begin{algorithmic}
  \small
  \State \textbf{Input:} An \el{} theory $T$
  \State \textbf{Output:} An extended theory $T'$\\
  
  \For{all $A \sqcap B \sqsubseteq E$ in the knowledge base} \\
    \[
    \inference {A \sqcap B \sqsubseteq E \quad \boldsymbol{A' \sqsubseteq A} \quad \boldsymbol{B' \sqsubseteq B} \quad \boldsymbol{E \sqsubseteq E'}}{A' \sqcap B' \sqsubseteq E'}
    \] \\
\EndFor
\For{all $A \sqsubseteq \exists r.B$ in the knowledge base} \\
    \[
    \inference {A \sqsubseteq \exists r.B \quad \boldsymbol{A' \sqsubseteq A} \quad \boldsymbol{B \sqsubseteq B'} \quad r \sqsubseteq r'}{A' \sqsubseteq \exists r'.B'} \quad 
    \inference {A \sqsubseteq \exists r.B \quad B \sqsubseteq \exists r'.E \quad r \circ r' \sqsubseteq s}{A \sqsubseteq \exists s.E}
    \] \\
\EndFor
\For{all $\exists r.A \sqsubseteq B$ in the knowledge base} \\
    \[
    \inference {\exists r.A \sqsubseteq B \quad \boldsymbol{A' \sqsubseteq A} \quad \boldsymbol{B \sqsubseteq B'} \quad r' \sqsubseteq r}{\exists r'.A' \sqsubseteq B'}
    \] \\
\EndFor
\For{all $A \sqcap B \sqsubseteq \bot$ in the knowledge base} \\
    \[
    \inference {A \sqcap B \sqsubseteq \bot \quad \boldsymbol{A' \sqsubseteq A} \quad \boldsymbol{B' \sqsubseteq B}}{A' \sqcap B' \sqsubseteq \bot} \quad \inference {A \sqcap B \sqsubseteq \bot}{A \sqcap B \sqsubseteq E}
    \] \\
\EndFor
\For{all $\exists r.A \sqsubseteq \bot$ in the knowledge base} \\
    \[
    \inference {\exists r.A \sqsubseteq \bot \quad \boldsymbol{A' \sqsubseteq A} \quad r' \sqsubseteq r}{\exists r'.A' \sqsubseteq \bot}
    \]
\EndFor \\
\label{algo}
\end{algorithmic}
\end{algorithm}

\section{Negative Sampling and Entailments} \label{negative_sampling}

In the case of knowledge base completion where the deductive closure
contains potentially many non-trivial entailed axioms, the random
sampling approach for negatives may lead to suboptimal learning since
some of the axioms treated as negatives may be entailed (and should
therefore be true in any model, in particular the one constructed by
the geometric embedding method). As an example, let us consider the
simple ontology consisting of two axioms: 
$A \sqcap B \sqsubseteq E$ and $F \sqsubseteq B$. For the $A \sqcap B \sqsubseteq E$
axiom, random negative sampling will sample $A \sqcap B \sqsubseteq E'$ 
where $E'$ is one of $A, B, E, F$. Since the knowledge base makes the axioms
$A \sqcap B \sqsubseteq A$, $A \sqcap B \sqsubseteq B$, and
$A \sqcap B \sqsubseteq E$ true, in 75\% of cases we will sample a
negative for this axiom that is actually true in each model.

\begin{algorithm}[h!]\captionsetup{labelfont={sc,bf}, labelsep=newline} 
  \caption{Additional entailed axioms. The input of the
      algorithm is the set of concept and/or role names, the output is
      the set of entailed axioms computed according to syntactic rules
      stated in the algorithm.}
\begin{algorithmic}
    \State \textbf{Input:} An \el{} theory $T$
  \State \textbf{Output:} An extended theory $T'$\\

    \For{all concepts $A, B, E, E'$ in the signature} \\
        \[
        \begin{split}
            \inference {}{A \sqcap \bot \sqsubseteq E} \quad \inference {B \sqsubseteq \bot}{A \sqcap B \sqsubseteq E} \quad \inference {E \sqsubseteq E'}{A \sqcap E \sqsubseteq E'} \quad \inference{A \sqcap B \sqsubseteq \bot}{A \sqcap B \sqsubseteq E} \\ \inference {A \sqsubseteq E \quad B \sqsubseteq E \quad A' \sqsubseteq A \quad B' \sqsubseteq B \quad E \sqsubseteq E'}{A' \sqcap B' \sqsubseteq E'} \quad \inference{A \sqsubseteq A'}{A \sqcap \top \sqsubseteq A'}
        \end{split}
        \] \\
    \EndFor
    \For{all roles $r$ and all concepts $B \neq \bot$ in the signature} \\
        \[
        \inference {}{\bot \sqsubseteq \exists r.B} \quad \inference {A \sqsubseteq \bot}{A \sqsubseteq \exists r.B}
        \] \\
    \EndFor
    \For{all roles $r$ and all concepts $A \neq \bot$ in the signature} \\
        \[
        \inference {}{\exists r.A \sqsubseteq \top}
        \] \\
    \EndFor \\
\end{algorithmic}
\end{algorithm}

We suggest to filter selected negatives during training based on the
deductive closure of the knowledge base: for each randomly generated
axiom to be used as negative, we check whether it is present in the
deductive closure and, if it is, we delete it.  \el reasoners such as
ELK~\cite{elk} compute subsumption hierarchies, i.e., all axioms of
the form $A \sqsubseteq B$ in the deductive closure, but not entailed
axioms for the other normal forms.  We use the inferences computed by
ELK (of the form $A \sqsubseteq B$ where $A$ and $B$ are concept names)
to design an algorithm that computes (a part of) the deductive closure
with respect to the \el normal forms. The algorithm implements a sound
but incomplete set of inference rules which can quickly generate a
partial deductive closure with respect to all normal forms. Algorithm
1 contains inference rules for deriving entailed axioms of type GCI1,
GCI2, GCI3, GCI1-BOT and GCI3-BOT from axioms explicitly represented
within a knowledge base; GCI0 and GCI0-BOT axioms are precomputed by
ELK. Algorithm 2 provides a set of additional rules depending on
arbitrary classes and roles represented within a knowledge base
after inferred axioms from Algorithm 1 are computed. The
  purpose of Algorithm 2 is to enrich the approximate deductive
  closure with axioms involving arbitrary roles and concepts or
  with axioms of new GCI type which may be missed by applying rules
  from Algorithm 1 since Algorithm 1 computes entailed axioms based on
  ontology axioms, concept hierarchy and role inclusions. 
  For example, consider an ontology consisting of two axioms:
  $A \sqsubseteq E$ and $B \sqsubseteq E$. Since no GCI1 axioms are
  present in the ontology, no GCI1 axioms will be entailed by
  Algorithm 1. Algorithm 2 enables inference of a GCI1 axiom
  $A \sqcap B \sqsubseteq E$. Another example is an ontology comprised
  of axioms $A \sqsubseteq B$ and $B \sqcap E \sqsubseteq F$. The
  axiom $A \sqcap \bot \sqsubseteq B$, which is entailed, cannot be
  inferred by applying rules from Algorithm 1: the concept $A$ is not
  a subclass of neither $B$ or $E$. Although we can use ELK or
similar reasoners to query for arbitrary entailed axioms, the
algorithms we propose have an advantage over this method since it does
not require the addition of a new concept to an ontology and
recomputing the concept hierarchy.  We show a detailed example of the
algorithm in Appendix~\ref{dc_example} based on the simple ontology
example introduced in Section~\ref{results}.

In the task of knowledge base completion with many non-trivial
entailed axioms, the deductive closure can also be used to modify the
evaluation metrics, or define novel evaluation metrics that
distinguish between entailed and non-entailed axioms. So far, ontology
embedding methods that have been applied to the task of knowledge base
completion have used evaluation measures that are taken from the task
of knowledge graph completion; in particular, they only evaluate
knowledge base completion using axioms that are ``novel'' and not
entailed. However, any entailed axiom will be true in all models of
the knowledge base, and therefore also in the geometric model that is
constructed by the embedding method.

We suggest to filter entailed axioms from training or test sets when the aim is to
predict ``novel'' (i.e., non-entailed) knowledge. The geometric embedding methods generate models making all entailed axioms true in all models.
It is expected that methods explicitly constructing models preferentially make entailed axioms true and rank them higher than non-entailed axioms. If the evaluation is based solely on non-entailed axioms, it will consider all similar inferred axioms false, and to avoid this, we may filter such axioms from the ranking list. The more axioms are filtered, the more entailed axioms are predicted
by a model.

\section{Experiments} \label{results}

\subsection{Datasets}

\subsubsection{Gene Ontology \& STRING Data} \label{go_string}

Following previous
works~\cite{kulmanov2019embeddings,peng2022description,jackermeier2023box}
we use common benchmarks for knowledge-base completion, in particular
a task that predicts protein--protein interactions (PPIs) based on the
functions of proteins. We also use the same data for the task of
protein function prediction. For these tasks we use two datasets, each
of them consists of the Gene Ontology (GO)~\cite{gene2015gene} with
all its axioms, protein--protein interactions (PPIs) and protein
function axioms extracted from the STRING
database~\cite{mering2003string}; each dataset focuses on only yeast
proteins. GO is formalized using OWL 2 EL~\cite{horrocksobo}.

For the PPI yeast network we use the built-in dataset {\tt
  PPIYeastDataset} available in the
mOWL~\cite{10.1093/bioinformatics/btac811} Python library (release
0.2.1) where axioms of interest are split randomly into train,
validation and test datasets in ratio 90:5:5 keeping pairs of
symmetric PPI axioms within the same dataset, and other axioms are
placed into the training part; validation and test sets are made up of
TBox axioms of type
$\{P_1\} \sqsubseteq \exists interacts\_with.\{P_2\}$ where $P_1, P_2$
are protein names. The GO version released on 2021-10-20 and the
STRING database version 11.5 were used. Alongside with the yeast
$interacts\_with$ dataset we collected the yeast $has\_function$
dataset organized in the same manner with validation and test parts
containing TBox axioms of type
$\{P\} \sqsubseteq \exists has\_function.\{GO\}$.  Based on the
information in the STRING database, in PPI yeast, the {\em
  interacts\_with} role is symmetric and the dataset is closed
against symmetric interactions. We normalize each train ontology using
the updated implementation of the jcel~\cite{mendez2012jcel} reasoner
\footnote{https://github.com/julianmendez/jcel/pull/12} where we take
into consideration newly generated concept and role names. Although
role inclusion axioms may be utilized within the $Box^2EL$ framework
we ignore them since neither {\it ELEmbeddings} nor {\it ELBE}
incorporate these types of axioms. Table in the appendix~\ref{go_string_stat} shows the number of GCIs of each type in the datasets and the number of concepts and roles after normalization. For more precise evaluation
of novel knowledge prediction we remove entailed axioms from the test
set for function prediction task based on the precomputed deductive
closure of the train ontology (see Section~\ref{negative_sampling}).

\subsubsection{Food ontology \& GALEN Ontology}

Food Ontology~\cite{dooley2018foodon} contains structured information
about foods formalized in $\mathcal{SRIQ}$ DL
expressivity~\cite{owl2vecstar} involving terms from
UBERON~\cite{mungall2012uberon}, NCBITaxon~\cite{federhen2012ncbi},
Plant Ontology~\cite{jaiswal2005plant}, and others. The
  GALEN ontology~\cite{rector1996galen} represents biomedical concepts
  related to anatomy, diseases, and others~\cite{rector2004patterns}.
For the Food Ontology, the data for subsumption prediction
was extracted from the case studies used to evaluate
OWL2Vec*~\cite{owl2vecstar}\footnote{https://github.com/KRR-Oxford/OWL2Vec-Star/tree/master/case\_studies};
the train part of the ontology was restricted to the
$\mathcal{EL}$ fragment and normalized using the
jcel~\cite{mendez2012jcel} reasoner. In case of GALEN
  ontology, subsumption axioms were randomly split in ratio 90:5:5
  among train, validation and test sets.  Since the normalization
procedure splits each complex axiom into a set of shorter axioms
including subsumptions between atomic concepts from the signature, it
may result in adding axioms represented in the validation or test part
of the ontology to the train part. To avoid this, we
filtered out such axioms from the original validation and test
datasets after the train ontology for subsumption prediction was
normalized. Additionally, as described in Section~\ref{go_string}, we
remove entailed axioms from the test dataset. Statistics about the
number of axioms of each GCI type, roles and classes can be found
in Appendix~\ref{food_stat} for the Food Ontology and in
  Appendix~\ref{galen_stat} for the GALEN ontology.

\subsection{Evaluation Scores and Metrics}

For GO \& STRING data, we predict GCI2 axioms of type
$\{P_1\} \sqsubseteq \exists interacts\_with.\{P_2\}$ or
$\{P\} \sqsubseteq \exists has\_function.\{GO\}$ depending on the
dataset. On Food Ontology and GALEN ontology, we predict
GCI0 axioms of type $A \sqsubseteq B$, $A$ and $B$ are arbitrary
classes from the signature. For each axiom type, we use the
corresponding loss expressions to score axioms. This is justified by
the fact that objective functions are measures of truth for each axiom
within constructed models.

The predictive performance is measured by the Hits@n metrics for
$n = 10, 100$, macro and micro mean rank, and the area under the
ROC curve (AUC ROC). For rank-based metrics, we calculate the score of $A \sqsubseteq \exists r.B$ or $A \sqsubseteq B$ 
for every class $A$ from the test set and for every $B$ from the set $\mathbf{C}$ of all
classes (or subclasses of a certain type, such as proteins or
functions for domain-specific cases) and determine the rank of a test
axiom $A \sqsubseteq \exists r.B$.  For macro mean rank and AUC ROC,
we consider all axioms from the test set; for micro metrics, we
compute corresponding class-specific metrics averaging them over all
classes in the signature:
\begin{align}
  micro\_MR_{A \sqsubseteq \exists r.B} &= Mean(MR_A(\{A \sqsubseteq \exists r.B, \thinspace B \in \mathbf{C}\})) \\
  micro\_MR_{A \sqsubseteq B} &= Mean(MR_A(\{A \sqsubseteq B, \thinspace B \in \mathbf{C}\})) \\
  micro\_AUC\_ROC_{A \sqsubseteq \exists r.B} &= Mean(AUC\_ROC_A(\{A \sqsubseteq \exists r.B, \thinspace B \in \mathbf{C}\})) \\
  micro\_AUC\_ROC_{A \sqsubseteq B} &= Mean(AUC\_ROC_A(\{A \sqsubseteq B, \thinspace B \in \mathbf{C}\}))
\end{align}

Additionally, we remove axioms represented in the train set or
deductive closures (see Section~\ref{negative_sampling}) to obtain corresponding
filtered metrics (FHits@n, FMR, FAUC). In related work
  focusing on knowledge graph completion or knowledge base completion
  tasks~\cite{bordes2013translating,wang2014knowledge,kulmanov2019embeddings,peng2022description},
  filtered metrics are computed by removing axioms presented within
  the train set from the list of all ranked axioms. This filtration is
  applied to eliminate statements learnt by a model during training
  phase which are therefore likely to have lower rank and to evaluate
  the predictive performance of a model in a more fair setting.

\subsection{Training Procedure}~\label{training_procedure}

All models are optimized with respect to the sum of individual GCI
losses (here we define the loss in most general case using all
positive and all negative losses):
\begin{align}
\begin{split}
    {\mathcal L} = l_{A \sqsubseteq B} + l_{A \sqcap B \sqsubseteq E} + l_{A \sqsubseteq \exists r.B} + l_{\exists r.A \sqsubseteq B} + l_{A \sqsubseteq \bot} + l_{A \sqcap B \sqsubseteq \bot} + l_{\exists r.A \sqsubseteq \bot} + \\ + l_{A \not\sqsubseteq B} + l_{A \sqcap B \not\sqsubseteq E} + l_{A \not\sqsubseteq \exists r.B} + l_{\exists r.A \not\sqsubseteq B} + l_{A \not\sqsubseteq \bot} + l_{A \sqcap B \not\sqsubseteq \bot} + l_{\exists r.A \not\sqsubseteq \bot}
\end{split}
\end{align}

All model architectures are built using the
mOWL~\cite{10.1093/bioinformatics/btac811} library on top of mOWL's
base models. All models were trained using the same fixed random seed.

All models are trained for 2,000 epochs for STRING \& GO datasets and
800 epochs for the Food Ontology and GALEN
datasets with batch size of 32,768. Training and
optimization is performed using Pytorch with Adam
optimizer~\cite{kingma2014adam} and ReduceLROnPlateau scheduler with
patience parameter $10$. We apply early stopping if validation loss
does not improve for $20$ epochs. For {\it ELEmbeddings},
hyperparameters are tuned using grid search over the following set:
margin $\gamma \in \{-0.1, -0.01, 0, 0.01, 0.1\}$, embedding dimension
$\{50, 100, 200, 400\}$, learning rate $\{0.01, 0.001, 0.0001\}$;
since none of our datasets contains unsatisfiable classes, we do not
tune the parameter $\varepsilon$ appearing in GCI0-BOT and GCI3-BOT
negative losses. For {\it ELBE}, grid search is performed over 60
randomly chosen subsets of the following hyperparameters: embedding
dimension $\{25, 50, 100, 200\}$, margin
$\{-0.1, -0.01, 0, 0.01, 0.1\}$,
$\varepsilon \in \{0.1, 0.01, 0.001\}$ (for experiments with all
negative losses involved), learning rate $\{0.01, 0.001,
0.0001\}$. The same strategy is applied to $Box^2EL$ models for
embedding dimension $\{25, 50, 100, 200\}$, margin
$\gamma \in \{-0.1, -0.01, 0, 0.01, 0.1\}$, $\delta \in \{1, 2, 4\}$,
$\varepsilon \in \{0.1, 0.01, 0.001\}$ (similarly, for experiments
with all negative losses involved), regularization factor
$\lambda \in \{0, 0.05, 0.1, 0.2\}$, and learning rate
$\{0.01, 0.001, 0.0001\}$. For experiments with negatives filtration
during training we use the same set of hyperparameters for random and
filtered mode of negative sampling. See Appendix~\ref{hyperparameters}
for details on optimal hyperparameters used.

\subsection{Results} \label{results}

We evaluate whether adding negative losses for all normal forms will
allow for the construction of a better model and improve the
performance in the task of knowledge base completion. We test the effect of the expanded negative sampling and negative losses first on a small ontology that can be
embedded and visualized in 2D space, and then on a larger application. We formulate and
add negative losses for all normal forms given by
equations~\ref{eq1}--\ref{box2el_eq6}.

\begin{figure}[!h]
\begin{subfigure}{0.49\textwidth}
    \centering
    \includegraphics[width=\textwidth]{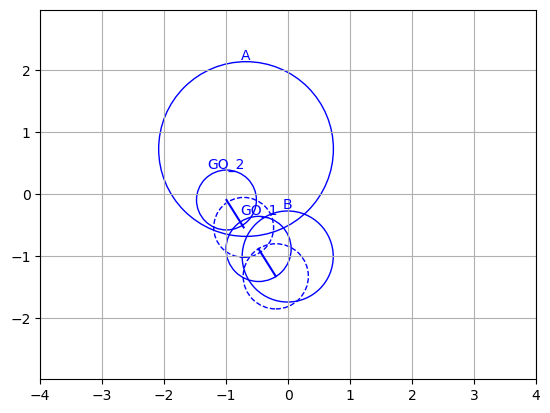}
    \caption{GCI2 negative loss}
    \label{fig:elem_toy}
\end{subfigure}
\hfill
\begin{subfigure}{0.49\textwidth}
    \centering
    \includegraphics[width=\textwidth]{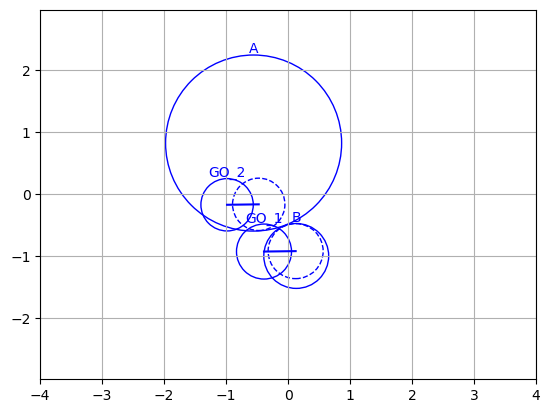}
    \caption{All negative losses}
    \label{fig:elem_losses_toy}
\end{subfigure}
\label{fig:elem_losses}
\caption{{\it ELEmbeddings} example. Dashed circles represent
  translated classes by role vector corresponding to
  $has\_function$ role. The normalized theory
    $\{\{GO_1\} \sqcap \{GO_2\} \sqsubseteq \bot, A \sqcap B
    \sqsubseteq \bot, \exists has\_function.\{GO_1\} \sqsubseteq B,
    \exists has\_function.\{GO_2\} \sqsubseteq A\}$ is better
    preserved when negative losses are incorporated to all normal
    forms (Figure b) rather than only to GCI2 normal form (Figure
    a).}
\end{figure}

First, we investigate a simple example corresponding to the task of
protein function prediction using the {\it ELEmbeddings} model. Let us
consider an ontology consisting of two axioms stating that there are
two disjoint functions $\{GO_1\}$ and $\{GO_2\}$, and proteins having
these functions are also disjoint:
$\{GO_1\} \sqcap \{GO_2\} \sqsubseteq \bot$,
$\exists has\_function.\{GO_1\} \sqcap \exists has\_function.\{GO_2\}
\sqsubseteq \bot$. After normalization, the last axiom is substituted
by the following three axioms: $A \sqcap B \sqsubseteq \bot$,
$\exists has\_function.\{GO_1\} \sqsubseteq B$,
$\exists has\_function.\{GO_2\} \sqsubseteq A$ where $A, B$ are new
concept names. To visualize the results, we embed these axioms in 2D
space. Figure~\ref{fig:elem_toy} shows the embedding generated with the
original {\it ELEmbeddings} model. Since there are no axioms of type
GCI2 represented within the knowledge base, the model learns without
any negative examples and demonstrates poor performance compared to
the model with incorporated negative losses for all normal forms as
demonstrated in Figure~\ref{fig:elem_losses_toy}.

Since we are interested in predicting not only axioms of type
$A \sqsubseteq \exists r.B$ for which negative sampling is used in the
original {\it ELEmbeddings}, {\it ELBE} and $Box^2EL$, we also examine
the effect of all negative losses utilization during training on Food
Ontology and GALEN ontology for subsumption prediction (see
Tables~\ref{table3} and \ref{table4}, respectively). We
find that the {\it ELEmbeddings} model does not improve on the Food
Ontology subsumption prediction task, but {\it ELBE} with additional
losses improves over the original model; $Box^2EL$ with additional
losses surpasses its version with just GCI2 negative loss in Hits@n
metrics. As for the results on GALEN ontology, we find that
  in case of all three models Hits@n metrics are improved when all
  negative losses are applied (except Hits@100 metric for {\it
    ELEmbeddings} model) indicating that in this particular case
  negative losses encourage models to predict more new axioms. Mean
  rank results are similarly better for {\it ELEmbeddings} and
  $Box^2EL$ models.

Additionally, we evaluate the performance on a standard benchmark set
for protein--protein interaction (PPI) prediction (see
Table~\ref{table1}). For this task, the test axioms are of the type
GCI2.  We observe that {\it ELEmbeddings} and {\it ELBE} with negative
losses for all normal forms integrated demonstrate superior
performance compared to their initial configurations in terms of
Hits@n metrics; it also allows $Box^2EL$ to lower ranks of test
axioms. Generally, for the task of PPI prediction, additional negative
sampling improves performance.

To summarize the above mentioned observations, we note that in some cases additional negative losses may decrease the ability of models to predict new
axioms and encourage models to predict entailed knowledge first (as,
e.g., in protein function prediction case) thus leading to
construction of a more accurate model of a theory. Since there is a
tradeoff between prediction of novel and entailed knowledge,
additional negative losses may demonstrate worse performance on novel
knowledge prediction.

\begin{figure}[!h] 
\begin{subfigure}{0.49\textwidth}
    \centering
    \includegraphics[width=\textwidth]{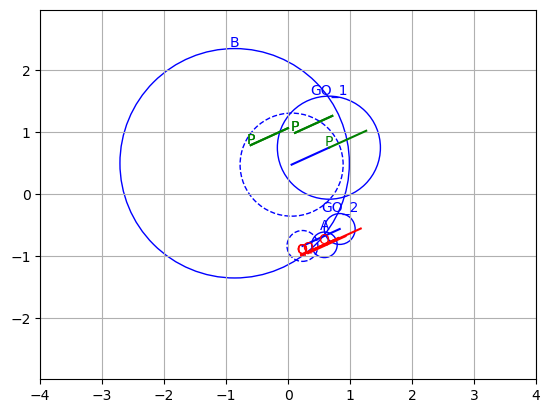}
    \caption{With random negatives}
    \label{fig:elem_losses_for_neg_toy}
\end{subfigure}
\hfill
\begin{subfigure}{0.49\textwidth}
    \centering
    \includegraphics[width=\textwidth]{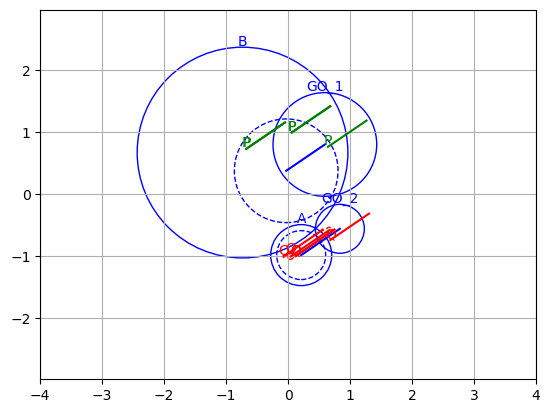}
    \caption{With filtered negatives}
    \label{fig:elem_losses_neg_filter_toy}
\end{subfigure}
\caption{{\it ELEmbeddings} example. Dashed circles represent
  translated classes by role vector corresponding to $has\_function$ role. `Red' classes represent proteins $\{Q_1\},
  \dots, \{Q_5\}$, `green' classes represent proteins $\{P_1\}, \dots,
  \{P_5\}$. Axioms $\{Q_i\} \sqsubseteq \exists
    has\_function.\{GO_2\}, i = 1, \dots, 5$ are better preserved when
    negatives are filtered based on precomputed deductive closure
    (Figure b) rather than when random negatives are sampled (Figure
    a). The same applies for the axiom $\exists has\_function.\{GO_2\} \sqsubseteq A$.}
\label{fig:elem_for_neg_filter}
\end{figure}

Using the example introduced above and
the {\it ELEmbeddings} embedding model, we demonstrate that negatives
filtration may be beneficial for constructing a model of a
theory. Apart from axioms mentioned earlier, i.e.,
$\{GO_1\} \sqcap \{GO_2\} \sqsubseteq \bot$,
$A \sqcap B \sqsubseteq \bot$,
$\exists has\_function.\{GO_1\} \sqsubseteq B$ and
$\exists has\_function.\{GO_2\} \sqsubseteq A$, we add 10 more axioms
about 5 proteins $\{P_1\}, \dots, \{P_5\}$ having function $\{GO_1\}$
(i.e.,
$\{P_i\} \sqsubseteq \exists has\_function.\{GO_1\}, \thickspace i =
1, \dots, 5$), and 5 proteins $\{Q_1\}, \dots, \{Q_5\}$ having
function $\{GO_2\}$ (i.e.,
$\{Q_i\} \sqsubseteq \exists has\_function.\{GO_2\}, \thickspace i =
1, \dots, 5$). Figure~\ref{fig:elem_for_neg_filter} shows the
constructed models with and without negatives filtering. We observe
that the model with filtered negatives provides faithful
representation of GCI3 axiom
$\exists has\_function.\{GO_2\} \sqsubseteq A$ and axioms introducing
proteins having function $\{GO_2\}$ as opposed to its counterpart with
random negatives: according to geometric interpretation, for GCI3 axioms $\exists r.A \sqsubseteq B$ to be faithfully
  represented, the $n$-ball interpreting the concept $A$ translated by $-r_0$ role $r$ vector should lie inside the $n$-ball interpreting
  the concept $B$. We see that this hold true on
  Figure~\ref{fig:elem_losses_neg_filter_toy} yet not on
  Figure~\ref{fig:elem_losses_for_neg_toy}.

Tables~\ref{table1}-- \ref{table4} show results in the
tasks of protein--protein interaction and subsumption prediction. We
find that excluding axioms in the deductive closure for negative
selection slightly improves or yields similar results. One possible
reason is that a randomly chosen axiom is very unlikely to be entailed
since very few axioms are entailed compared to all possible axioms to
choose from.

\begin{figure}[!h]
\begin{subfigure}{0.49\textwidth}
    \centering
    \includegraphics[width=\textwidth]{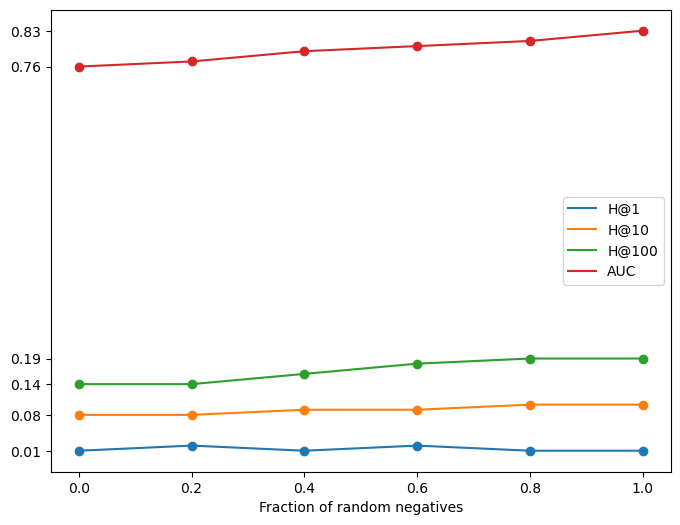}
    \caption{H@1, H@10, H@100 and ROC AUC}
    \label{fig:random_neg_fraction_1}
\end{subfigure}
\hfill
\begin{subfigure}{0.49\textwidth}
    \centering
    \includegraphics[width=\textwidth]{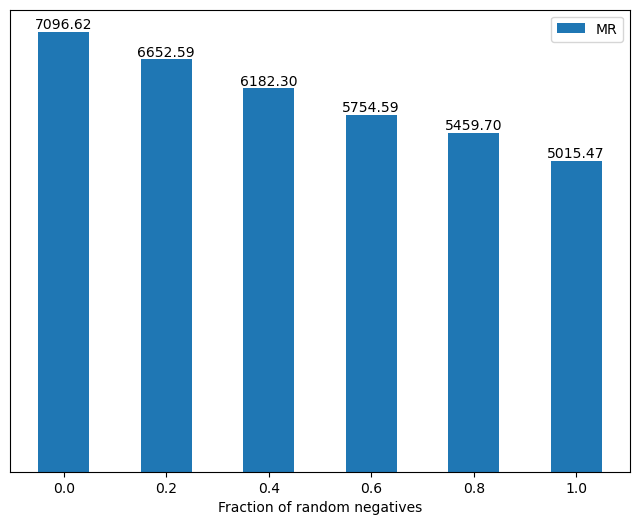}
    \caption{macro\_MR}
    \label{fig:random_neg_fraction_2}
\end{subfigure}
\caption{Metrics reported for biased fraction of random
    negatives combined with entailed axioms from the precomputed
    deductive closure.}
\label{fig:neg_fraction}
\end{figure}

Because the chance of selecting an entailed axiom as a negative
depends on the knowledge base on which the embedding method is
applied, we perform additional experiments on Food Ontology with {\it
  ELEmbeddings} model where we bias the selection of negatives; we
chose between 100\% negatives to 0\% negatives from the entailed
axioms.  We find that reducing the number of entailed axioms from the
negatives has an effect to improve performance and the effect
increases the more axioms would be chosen from the entailed ones (see
Figure~\ref{fig:neg_fraction}).

We compute filtered metrics for the protein function and subsumption prediction tasks. Both of them account for entailed axioms prediction since if, e.g., $A \sqsubseteq B$ is being predicted then first models may predict axioms of type $A \sqsubseteq B'$ 
where $B'$ is any superclass of $B$; the same is true for function prediction
axioms $\{P\} \sqsubseteq \exists has\_function.\{GO\}$ and all
superclasses $\{GO'\}$ of $\{GO\}$ class. Note that the
protein--protein interaction prediction task is not tailored for
evaluation using deductive closures of the train or test set: for each
protein $\{P\}$ its subclasses include only $\bot$ and superclasses
include only $\top$. As a result, the only inferred axioms will be of
type $\bot \sqsubseteq \exists interacts\_with.\{P\}$, $\{P_1\}
\sqsubseteq \exists interacts\_with.\{P_2\}$ or $\{P\} \sqsubseteq
\exists interacts\_with.\top$, and filtered metrics may be computed
only with respect to the train part of the ontology. For
  this reason we do not report filtered metrics for protein--protein
  interaction prediction task (Table~\ref{table1}).

For function prediction and subsumption prediction, we employ
filtration of metrics based on the deductive closure of the train set
and of the test set. Tables~\ref{table3}, \ref{table4}
and~\ref{table2} contain results for subsumption prediction on Food
Ontology, subsumption prediction on GALEN ontology and
function prediction on GO, respectively.

Our findings suggest that the baseline {\it ELEmbeddings} predicts
primarily entailed axioms of GCI2 type, yet for GCI0 on Food
  Ontology the model predicts ``novel'' knowledge first whereas the
model modifications with additional negative losses and negatives
filtration derive entailed knowledge in the first place. For
  the GALEN ontology, however, the situation is similar to the protein
  function prediction case, i.e., novel knowledge is predicted in the
  first place for modifications with additional negative losses and
  negatives filtration. This may indicate model construction where
  many classes overlap or ``collapse'' for all negative losses and
  negatives filtering case since the GALEN ontology does not contain
  disjointness axioms and consequently no classes will be separated by
  the model. The same holds for {\it ELBE} and $Box^2EL$ models.
Losses for all normal forms and negatives filtering during training
aid {\it ELBE} and $Box^2EL$ to construct model-generated embeddings
which first predict logically inferred knowledge and then non-entailed
axioms of type GCI2 or GCI0 (on Food Ontology),
respectively. The results indicate that models with all types of valid
negatives in most cases explicitly construct models.

\section{Discussion}

We evaluated properties of {\it ELEmbeddings}, {\it ELBE} and
$Box^2EL$, ontology embedding methods that aims to generate a model of
an \el theory; the properties we evaluate hold similarly for other
ontology embedding methods that construct models of \el
theories. While we demonstrate several improvements over the original
model, we can also draw some general conclusions about ontology
embedding methods and their evaluation. Knowledge base completion is
the task of predicting axioms that should be added to a knowledge
base; this task is adapted from knowledge graph completion where
triples are added to a knowledge graph. The way both tasks are
evaluated is by removing some statements (axioms or triples) from the
knowledge base, and evaluating whether these axioms or triples can be
recovered by the embedding method. This evaluation approach is
adequate for knowledge graphs which do not give rise to many
entailments. However, knowledge bases give rise to potentially many
non-trivial entailments that need to be considered in the
evaluation. In particular, embedding methods that aim to generate a
model of a knowledge base will first generate entailed axioms (because
entailed axioms are true in all models); these methods perform
knowledge base completion as a generalization of generating the model
where either other statements may be true, or they may be
approximately true in the generated structure. This has two
consequences: the evaluation procedure needs to account for this; and
the model needs to be sufficiently rich to allow useful predictions.

We have introduced a method to compute the deductive closure of \el
know\-ledge bases; this method relies on an automated reasoner and is
sound. 
We use all the axioms in the deductive closure
as positive axioms to be predicted when evaluating knowledge base
completion, to account for methods that treat knowledge base
completion as a generalization of constructing a model and testing for
truth in this model. We find that some models (e.g., modified box-based models using valid negatives of all types) can predict entailed axioms well, some (e.g., the original $Box^2EL$ model) preferentially predict ``novel'', non-entailed axioms;
these methods solve subtly different problems (either generalizing
construction of a model, or specifically predicting novel non-entailed
axioms). 
We also modify the evaluation procedure to account for the
inclusion of entailed axioms as positives; however, the evaluation
measures are still based on ranking individual axioms and do not
account for semantic similarity. For example, if during testing, the
correct axiom to predict is $A \sqsubseteq \exists r.B$ but the predicted axiom is $A \sqsubseteq \exists r.E$, the prediction may be
considered to be ``more correct'' if $B \sqsubseteq E$ was in the knowledge base than if $B \sqcap E \sqsubseteq \bot$ 
was in the knowledge base. Novel evaluation metrics need to be designed to
account for this phenomenon, similarly to ontology-based evaluation
measures used in life sciences \cite{radivojac2013}. It is also
important to expand the set of benchmark sets for knowledge base
completion.

Use of the deductive closure is not only useful in evaluation but also
when selecting negatives. In formal knowledge bases, there are at
least two ways in which negatives for axioms can be chosen: they are
either non-entailed axioms, or they are axioms whose negation is
entailed. However, in no case should entailed axioms be considered as
negatives; we demonstrate that filtering entailed axioms from selected
negatives during training improves the performance of the embedding
method consistently in knowledge base completion (and, obviously, more
so when entailed axioms are considered as positives during
evaluation).

While we only report our experiments with {\it ELEmbeddings}, {\it
  ELBE} and $Box^2EL$, our findings, in particular about the
evaluation and use of deductive closure, are applicable to other
geometric ontology embedding methods. As ontology embedding methods
are increasingly applied in knowledge-enhanced learning and other
tasks that utilize some form of approximate computation of
entailments, our results can also serve to improve the applications of
ontology embeddings.

\begin{table}[!h] 
  \caption{Protein function prediction experiments on yeast
    proteins. `l' corresponds to all negative losses, `l+n' means a
    model was trained using all negative losses and negatives
    filtering. For each model we report non-filtered metrics (NF) and
    filtered metrics with respect to the deductive closure of the
    train and the test set combined together (F). For
      macro\_MR and micro\_MR we additionally report the difference
      between filtered and non-filtered metrics (NF-F) to check how
      much of entailed knowledge is predicted on average. Values in
    {\bf bold} indicate best metrics.} \label{table2}
  \centering
  \resizebox{\textwidth}{!}{%
    \begin{tabular}{| C{1.5cm} || C{0.5cm} | C{0.5cm} ||
          C{0.5cm} | C{0.5cm} || C{0.71cm} | C{0.71cm} | C{0.61cm} ||
          C{0.71cm} | C{0.71cm} | C{0.61cm} || C{0.5cm} | C{0.5cm} ||
          C{0.5cm} | C{0.5cm} |} \hline
                \multirow{2}{*}{Model} & \multicolumn{2}{c||}{H@10} & \multicolumn{2}{c||}{H@100} & \multicolumn{3}{c||}{macro\_MR} & \multicolumn{3}{c||}{micro\_MR} & \multicolumn{2}{c||}{macro\_AUC} & \multicolumn{2}{c|}{micro\_AUC} \\
                \cline{2-15}
                                       & NF & F & NF & F & NF & F & NF-F& NF & F & NF-F & NF & F & NF & F \\
                \hhline{|===============|}
                ELEm & {\bf 0.01} & {\bf 0.01} & {\bf 0.03} & {\bf 0.03} & 21198 & 21150 & {\bf 48} & 21165 & 21118 & {\bf 47} & 0.62 & 0.62 & 0.63 & 0.63 \\
                \hline
                ELEm+l & 0.00 & 0.00 & {\bf 0.03} & {\bf 0.03} & 9603 & 9575 & 28 & 9449 & 9423 & 26 & {\bf 0.83} & {\bf 0.83} & {\bf 0.84} & {\bf 0.84} \\
                \hline
                ELEm+l+n & 0.00 & 0.00 & {\bf 0.03} & {\bf 0.03} & {\bf 9488} & {\bf 9460} & 28 & {\bf 9334} & {\bf 9307} & 27 & {\bf 0.83} & {\bf 0.83} & {\bf 0.84} & {\bf 0.84} \\
                \hhline{|===============|}
                ELBE & {\bf 0.03} & {\bf 0.03} & {\bf 0.24} & {\bf 0.24} & {\bf 4229} & {\bf 4209} & 20 & {\bf 4156} & {\bf 4137} & 19 & {\bf 0.92} & {\bf 0.92} & {\bf 0.93} & {\bf 0.93} \\
                \hline
                ELBE+l & 0.00 & 0.00 & 0.01 & 0.01 & 12920 & 12865 & {\bf 55} & 12797 & 12745 & 52 & 0.77 & 0.77 & 0.78 & 0.78 \\
                \hline
                ELBE+l+n & 0.00 & 0.00 & 0.01 & 0.01 & 12900 & 12845 & {\bf 55} & 12772 & 12719 & {\bf 53} & 0.77 & 0.77 & 0.78 & 0.78 \\
                \hhline{|===============|}
                $Box^2EL$ & {\bf 0.28} & {\bf 0.31} & {\bf 0.55} & {\bf 0.55} & {\bf 1988} & {\bf 1979} & {\bf 9} & {\bf 1988} & {\bf 1980} & {\bf 8} & {\bf 0.96} & {\bf 0.96} & {\bf 0.97} & {\bf 0.97} \\
                \hline
                $Box^2EL$+l & 0.24 & 0.27 & 0.54 & {\bf 0.55} & 2129 & 2120 & {\bf 9} & 2099 & 2091 & {\bf 8} & {\bf 0.96} & {\bf 0.96} & {\bf 0.97} & {\bf 0.97} \\
                \hline
                $Box^2EL$+l+n & 0.24 & 0.27 & 0.54 & {\bf 0.55} & 2161 & 2152 & {\bf 9} & 2147 & 2139 & {\bf 8} & {\bf 0.96} & {\bf 0.96} & 0.96 & 0.96 \\
                \hline
\end{tabular}}
\end{table}

\begin{table}[!h]
  \caption{Protein--protein interaction prediction experiments on
    yeast proteins. `l' corresponds to all negative losses, `l+n'
    means a model was trained using all negative losses and negatives
    filtering. Non-filtered metrics are reported. Values in {\bf bold}
    indicate best metrics.} \label{table1}
  \centering
  \begin{tabular}{| C{1.5cm} | C{0.8cm} | C{0.9cm} |
        C{1.3cm} | C{1.3cm} | C{1.5cm} | C{1.4cm} |} \hline
              Model & H@10 & H@100 & macro\_MR & micro\_MR & macro\_AUC & micro\_AUC \\
              \hline
              ELEm & 0.05 & 0.31 & 599.21 & 701.57 & 0.90 & 0.90 \\
              ELEm+l & {\bf 0.06} & 0.35 & 532.93 & 681.02 & {\bf 0.91} & 0.90 \\
              ELEm+l+n & {\bf 0.06} & {\bf 0.37} & {\bf 519.62} & {\bf 671.19} & {\bf 0.91} & {\bf 0.91} \\
              \hline
              ELBE & 0.07 & 0.37 & {\bf 829.86} & {\bf 1123.47} & {\bf 0.91} & {\bf 0.89} \\
              ELBE+l & {\bf 0.08} & {\bf 0.40} & 984.92 & 1259.54 & 0.84 & 0.82 \\
              ELBE+l+n & {\bf 0.08} & {\bf 0.40} & 984.18 & 1281.20 & 0.84 & 0.82 \\
              \hline
              $Box^2EL$ & {\bf 0.05} & 0.57 & 215.07 & 287.16 & 0.96 & {\bf 0.96} \\
              $Box^2EL$+l & {\bf 0.05} & 0.57 & 200.85 & {\bf 250.17} & {\bf 0.97} & {\bf 0.96} \\
              $Box^2EL$+l+n & {\bf 0.05} & {\bf 0.58} & {\bf 197.73} & 250.47 & {\bf 0.97} & {\bf 0.96} \\
              \hline
\end{tabular}
\end{table}

\begin{table}[!h] 
\caption{Subsumption prediction experiments on Food Ontology. `l' corresponds to all negative losses, `l+n' means a model was trained using all negative losses and negatives filtering. For each model we report non-filtered metrics (NF) and filtered metrics with respect to the deductive closure of the train and the test set combined together (F). For macro\_MR and micro\_MR we additionally report the difference between filtered and non-filtered metrics (NF-F) to check how much of entailed knowledge is predicted on average. Values in {\bf bold} indicate best metrics.} \label{table3}
\centering
\resizebox{\textwidth}{!}{%
\begin{tabular}{| C{1.5cm} || C{0.5cm} | C{0.5cm} || C{0.5cm} | C{0.5cm} || C{0.71cm} | C{0.71cm} | C{0.61cm} || C{0.71cm} | C{0.71cm} | C{0.61cm} || C{0.5cm} | C{0.5cm} || C{0.5cm} | C{0.5cm} |}
\hline
\multirow{2}{*}{Model} & \multicolumn{2}{c||}{H@10} & \multicolumn{2}{c||}{H@100} & \multicolumn{3}{c||}{macro\_MR} & \multicolumn{3}{c||}{micro\_MR} & \multicolumn{2}{c||}{macro\_AUC} & \multicolumn{2}{c|}{micro\_AUC} \\
\cline{2-15}
& NF & F & NF & F & NF & F & NF-F & NF & F & NF-F & NF & F & NF & F \\
\hhline{|===============|}
ELEm & {\bf 0.12} & {\bf 0.12} & {\bf 0.21} & {\bf 0.21} & {\bf 4659} & {\bf 4656} & {\bf 3} & {\bf 4662} & {\bf 4659} & {\bf 3} & {\bf 0.84} & {\bf 0.84} & {\bf 0.84} & {\bf 0.84} \\
\hline
ELEm+l & 0.10 & 0.11 & 0.19 & 0.19 & 5015 & 5013 & 2 & 5020 & 5017 & {\bf 3} & 0.83 & 0.83 & 0.83 & 0.83 \\
\hline
ELEm+l+n & 0.10 & 0.11 & 0.19 & 0.19 & 5022 & 5019 & {\bf 3} & 5027 & 5024 & {\bf 3} & 0.83 & 0.83 & 0.83 & 0.83 \\
\hhline{|===============|}
ELBE & 0.01 & 0.01 & 0.09 & 0.09 & 6695 & 6692 & {\bf 3} & 6688 & 6686 & 2 & 0.77 & 0.77 & 0.77 & 0.77 \\
\hline
ELBE+l & {\bf 0.04} & {\bf 0.04} & {\bf 0.14} & {\bf 0.14} & 5428 & 5426 & 2 & 5412 & 5409 & {\bf 3} & {\bf 0.81} & {\bf 0.81} & {\bf 0.82} & {\bf 0.82} \\
\hline
ELBE+l+n & {\bf 0.04} & {\bf 0.04} & {\bf 0.14} & {\bf 0.14} & {\bf 5427} & {\bf 5424} & {\bf 3} & {\bf 5410} & {\bf 5408} & 2 & {\bf 0.81} & {\bf 0.81} & {\bf 0.82} & {\bf 0.82} \\
\hhline{|===============|}
$Box^2EL$ & 0.01 & 0.01 & 0.10 & 0.10 & {\bf 3900} & {\bf 3898} & 2 & {\bf 3877} & {\bf 3874} & {\bf 3} & {\bf 0.87} & {\bf 0.87} & {\bf 0.87} & {\bf 0.87} \\
\hline
$Box^2EL$+l & 0.04 & 0.04 & 0.13 & 0.13 & 7550 & 7547 & {\bf 3} & 7555 & 7553 & 2 & 0.74 & 0.74 & 0.74 & 0.74 \\
\hline
$Box^2EL$+l+n & {\bf 0.05} & {\bf 0.05} & {\bf 0.14} & {\bf 0.14} & 6865 & 6862 & {\bf 3} & 6869 & 6866 & {\bf 3} & 0.76 & 0.76 & 0.77 & 0.77 \\
\hline
\end{tabular}}
\end{table}

\section*{Acknowledgements}
This work has been supported by funding from King Abdullah University
of Science and Technology (KAUST) Office of Sponsored Research (OSR)
under Award No. URF/1/4355-01-01, URF/1/4675-01-01, URF/1/4697-01-01,
URF/1/5041-01-01, and REI/1/5334-01-01.  This work was supported by
the SDAIA--KAUST Center of Excellence in Data Science and Artificial
Intelligence (SDAIA--KAUST AI), by funding from King Abdullah
University of Science and Technology (KAUST) -- KAUST Center of
Excellence for Smart Health (KCSH) under award number 5932, and by
funding from King Abdullah University of Science and Technology
(KAUST) -- KAUST Center of Excellence for Generative AI under award
number 5940.  We acknowledge support from the KAUST Supercomputing
Laboratory.

\begin{table}[!h] 
\caption{Subsumption prediction experiments on GALEN
    Ontology. `l' corresponds to all negative losses, `l+n' means a
    model was trained using all negative losses and negatives
    filtering. For each model we report non-filtered metrics (NF) and
    filtered metrics with respect to the deductive closure of the
    train and the test set combined together (F). For macro\_MR and
    micro\_MR we additionally report the difference between filtered
    and non-filtered metrics (NF-F) to check how much of entailed
    knowledge is predicted on average. Values in {\bf bold} indicate
    best metrics.} \label{table4}
\centering
\resizebox{\textwidth}{!}{%
\begin{tabular}{| C{1.5cm} || C{0.5cm} | C{0.5cm} || C{0.5cm} | C{0.5cm} || C{0.71cm} | C{0.71cm} | C{0.61cm} || C{0.71cm} | C{0.71cm} | C{0.61cm} || C{0.5cm} | C{0.5cm} || C{0.5cm} | C{0.5cm} |}
\hline
\multirow{2}{*}{Model} & \multicolumn{2}{c||}{H@10} & \multicolumn{2}{c||}{H@100} & \multicolumn{3}{c||}{macro\_MR} & \multicolumn{3}{c||}{micro\_MR} & \multicolumn{2}{c||}{macro\_AUC} & \multicolumn{2}{c|}{micro\_AUC} \\
\cline{2-15}
& NF & F & NF & F & NF & F & NF-F & NF & F & NF-F & NF & F & NF & F \\
\hhline{|===============|}
ELEm & 0.15 & 0.16 & {\bf 0.35} & {\bf 0.35} & 9106 & 9105 & 1 & 9106 & 9105 & 1 & {\bf 0.82} & {\bf 0.82} & {\bf 0.82} & {\bf 0.82} \\
\hline
ELEm+l & {\bf 0.21} & {\bf 0.22} & 0.34 & 0.34 & {\bf 8977} & {\bf 8976} & 1 & {\bf 8977} & {\bf 8976} & 1 & {\bf 0.82} & {\bf 0.82} & {\bf 0.82} & {\bf 0.82} \\
\hline
ELEm+l+n & {\bf 0.21} & {\bf 0.22} & 0.33 & 0.34 & 9005 & 9003 & {\bf 2} & 9005 & 9003 & {\bf 2} & {\bf 0.82} & {\bf 0.82} & {\bf 0.82} & {\bf 0.82} \\
\hhline{|===============|}
ELBE & 0.08 & 0.08 & 0.22 & 0.22 & {\bf 11236} & {\bf 11234} & 2 & {\bf 11236} & {\bf 11234} & 2 & {\bf 0.77} & {\bf 0.78} & {\bf 0.77} & {\bf 0.77} \\
\hline
ELBE+l & {\bf 0.13} & {\bf 0.13} & {\bf 0.34} & {\bf 0.34} & 11884 & 11882 & 2 & 11884 & 11882 & 2 & 0.76 & 0.76 & 0.76 & 0.76 \\
\hline
ELBE+l+n & 0.12 & 0.12 & {\bf 0.34} & {\bf 0.34} & 11720 & 11717 & {\bf 3} & 11720 & 11717 & {\bf 3} & {\bf 0.77} & 0.77 & 0.76 & 0.76 \\
\hhline{|===============|}
$Box^2EL$ & 0.14 & 0.14 & 0.31 & 0.31 & 11724 & 11721 & {\bf 3} & 11724 & 11721 & {\bf 3} & {\bf 0.77} & {\bf 0.77} & 0.76 & 0.76 \\
\hline
$Box^2EL$+l & {\bf 0.16} & {\bf 0.16} & {\bf 0.37} & {\bf 0.37} & {\bf 11371} & {\bf 11369} & 2 & {\bf 11371} & {\bf 11369} & 2 & {\bf 0.77} & {\bf 0.77} & {\bf 0.77} & {\bf 0.77} \\
\hline
$Box^2EL$+l+n & {\bf 0.16} & {\bf 0.16} & {\bf 0.37} & {\bf 0.37} & 11378 & 11376 & 2 & 11378 & 11376 & 2 & {\bf 0.77} & {\bf 0.77} & {\bf 0.77} & {\bf 0.77} \\
\hline
\end{tabular}}
\end{table}

\nocite{*}
\bibliographystyle{ios1}           
\bibliography{bibliography}        

%

\appendix

\section{GO \& STRING data Statistics, Train Part} \label{go_string_stat}

\begin{table}[!h]
\centering
\begin{tabular}{| C{1cm} | C{0.8cm} | C{0.8cm} |  C{1cm} | C{0.8cm} | C{1.4cm} | C{1.4cm} | C{1.4cm} | C{0.9cm} | C{1.1cm} | C{1.3cm} | C{0.3cm} |}
\hline
Dataset & GCI0 & GCI1 & GCI2 & GCI3 & GCI0\_BOT & GCI1\_BOT & GCI3\_BOT & Classes & Roles & Test axioms \\
\hline
Yeast iw & 81,068 & 11,825 & 269,567 & 11,823 & 0 & 31 & 0 & 61,846 & 16 & 12,040 \\
\hline
Yeast hf & 81,068 & 11,825 & 290,433 & 11,823 & 0 & 31 & 0 & 61,850 & 16 & 1,530 \\
\hline
\end{tabular}
\end{table}

\section{Food Ontology Statistics, Train Part} \label{food_stat}
\begin{table}[!h]
\centering
\begin{tabular}{| C{0.8cm} | C{0.8cm} |  C{1cm} | C{0.8cm} | C{1.4cm} | C{1.4cm} | C{1.4cm} | C{0.9cm} | C{1.1cm} | C{1.3cm} | C{0.3cm} |}
\hline
GCI0 & GCI1 & GCI2 & GCI3 & GCI0\_BOT & GCI1\_BOT & GCI3\_BOT & Classes & Roles & Test axioms \\
\hline
21,795 & 1,267 & 10,719 & 897 & 0 & 495 & 0 & 24,969 & 43 & 5,752 \\
\hline
\end{tabular}
\end{table}

\section{GALEN Ontology Statistics, Train Part} \label{galen_stat}
\begin{table}[!h]
\centering
\begin{tabular}{| C{0.8cm} | C{0.8cm} |  C{1cm} | C{0.8cm} | C{1.4cm} | C{1.4cm} | C{1.4cm} | C{0.9cm} | C{1.1cm} | C{1.3cm} | C{0.3cm} |}
\hline
GCI0 & GCI1 & GCI2 & GCI3 & GCI0\_BOT & GCI1\_BOT & GCI3\_BOT & Classes & Roles & Test axioms \\
\hline
27,339 & 15,613 & 29,618 & 15,615 & 0 & 0 & 0 & 49,223 & 888 & 667 \\
\hline
\end{tabular}
\end{table}

\newpage

\section{Hyperparameters} \label{hyperparameters}

\begin{table}[!htp]
\centering
\begin{tabular}{| C{1cm} | C{1.2cm} | C{0.5cm} | C{0.8cm} |  C{0.6cm} | C{0.7cm} | C{0.2cm} | C{0.5cm} |}
\hline
Dataset & Model & dim & lr & $\gamma$ & $\epsilon$ & $\delta$ & $\lambda$ \\
\hline
\multirow{6}{=}{Yeast iw} & ELEm & 100 & 0.0001 & -0.10 & & & \\
\cline{2-8}
& ELEm+l & 50 & 0.0001 & 0.00 & & & \\
\cline{2-8}
& ELBE & 200 & 0.0001 & 0.00 & & & \\
\cline{2-8}
& ELBE+l & 200 & 0.0100 & 0.00 & 0.001 & & \\
\cline{2-8}
& $Box^2EL$ & 200 & 0.0010 & 0.01 & & 1 & 0.05 \\
\cline{2-8}
& $Box^2EL$+l & 200 & 0.0010 & 0.01 & 0.010 & 2 & 0.05 \\
\hline
\hline
\multirow{6}{=}{Yeast hf} & ELEm & 200 & 0.0001 & 0.01 & & & \\
\cline{2-8}
& ELEm+l & 50 & 0.0001 & -0.10 & & & \\
\cline{2-8}
& ELBE & 200 & 0.0001 & 0.10 & & & \\
\cline{2-8}
& ELBE+l & 200 & 0.0001 & 0.10 & 0.010 & & \\
\cline{2-8}
& $Box^2EL$ & 200 & 0.0100 & 0.10 & & 4 & 0.20 \\
\cline{2-8}
& $Box^2EL$+l & 200 & 0.0100 & 0.10 & 0.010 & 4 & 0.05 \\
\hline
\hline
\multirow{6}{=}{FoodOn} & ELEm & 400 & 0.0010 & -0.10 & & & \\
\cline{2-8}
& ELEm+l & 400 & 0.0010 & -0.10 & & & \\
\cline{2-8}
& ELBE & 200 & 0.0100 & 0.10 & & & \\
\cline{2-8}
& ELBE+l & 200 & 0.0100 & -0.01 & 0.001 & & \\
\cline{2-8}
& $Box^2EL$ & 100 & 0.0100 & 0.10 & & 1 & 0.20 \\
\cline{2-8}
& $Box^2EL$+l & 200 & 0.0010 & 0.10 & 0.010 & 4 & 0.10 \\
\hline
\hline
\multirow{6}{=}{GALEN} & ELEm & 400 & 0.0010 & -0.10 & & & \\
\cline{2-8}
& ELEm+l & 400 & 0.0010 & -0.01 & & & \\
\cline{2-8}	
& ELBE & 100 & 0.0010 & 0.10 & & & \\
\cline{2-8}
& ELBE+l & 200 & 0.0010 & 0.01 & 0.010 & & \\
\cline{2-8}
& $Box^2EL$ & 200 & 0.0010 & 0.00 & & 4 & 0.05 \\
\cline{2-8}
& $Box^2EL$+l & 200 & 0.0100 & 0.00 & 0.100 & 1 & 0.05 \\
\hline
\end{tabular}
\end{table}

\section{Deductive Closure Computation Example} \label{dc_example}

Let us add two more axioms to the simple ontology example from
Section~\ref{results} about proteins $\{P\}$ and $\{Q\}$
having functions $\{GO_1\}$ and $\{GO_2\}$, respectively. ELK will
infer the following class hierarchy:

\begin{center}
\renewcommand{\arraystretch}{1.5}
\begin{tabular}{c|l}
$E$ & Concepts $F$ where $E \sqsubseteq F$ \\
\hline
$\bot$ & $\bot$, $\{P\}$, $\{Q\}$, $A$, $B$, $\{GO_1\}$, $\{GO_2\}$, $\top$ \\
$\{P\}$ & $\{P\}$, $B$, $\top$ \\
$\{Q\}$ & $\{Q\}$, $A$, $\top$ \\
$A$ & $A$, $\top$ \\
$B$ & $B$, $\top$ \\
$\{GO_1\}$ & $\{GO_1\}$, $\top$ \\
$\{GO_2\}$ & $\{GO_2\}$, $\top$ \\
$\top$ & $\top$
\end{tabular}
\end{center}

For GCI2 axioms $\{P\} \sqsubseteq \exists has\_function.\{GO_1\}$ and $\{Q\} \sqsubseteq \exists has\_function.\{GO_2\}$ the algorithm will output

\begin{center}
\renewcommand{\arraystretch}{1.5}
\begin{tabular}{c|l}
$E$ & Concepts $F \neq \bot$ where $E \sqsubseteq \exists has\_function.F$ \\
\hline
$\bot$ & $\{P\}$, $\{Q\}$, $A$, $B$, $\{GO_1\}$, $\{GO_2\}$, $\top$ \\
$\{P\}$ & $\{GO_1\}$, $\top$ \\
$\{Q\}$ & $\{GO_2\}$, $\top$ \\
\end{tabular}
\end{center}

For GCI3 axioms $\exists has\_function.\{GO_1\} \sqsubseteq B$ and $\exists has\_function.\{GO_2\} \sqsubseteq A$ the algorithm will infer 

\begin{center}
\renewcommand{\arraystretch}{1.5}
\begin{tabular}{c|l}
$E \neq \bot$ & Concepts $F$ where $\exists has\_function.E \sqsubseteq F$ \\
\hline
$\{P\}$ & $\top$ \\
$\{Q\}$ & $\top$ \\
$A$ & $\top$ \\
$B$ & $\top$ \\
$\{GO_1\}$ & $B$, $\top$ \\
$\{GO_2\}$ & $A$, $\top$ \\
$\top$ & $\top$ \\
\end{tabular}
\end{center}

In this small protein function prediction example there are two
disjointness axioms: $A \sqcap B \sqsubseteq \bot$ and
$\{GO_1\} \sqcap \{GO_2\} \sqsubseteq \bot$. Taking into consideration
the concept hierarchy and inference rules from part 2 the algorithm
will infer the following GCI1 and GCI1\_BOT axioms:

\begin{center}
\renewcommand{\arraystretch}{1.5}
\begin{adjustbox}{height=0.5\textheight}
  \begin{tabular}{c|c|l}
    $E$ & $F$ & Subsumptions $G$ where $E \sqcap F \sqsubseteq G$ \\
    \hline
    \multirow{8}{*}{$\bot$} & $\bot$ & $\bot$, $\{P\}$, $\{Q\}$, $A$, $B$, $\{GO_1\}$, $\{GO_2\}$, $\top$ \\
        & $\{P\}$ & $\bot$, $\{P\}$, $\{Q\}$, $A$, $B$, $\{GO_1\}$, $\{GO_2\}$, $\top$ \\
        & $\{Q\}$ & $\bot$, $\{P\}$, $\{Q\}$, $A$, $B$, $\{GO_1\}$, $\{GO_2\}$, $\top$ \\
        & $A$ & $\bot$, $\{P\}$, $\{Q\}$, $A$, $B$, $\{GO_1\}$, $\{GO_2\}$, $\top$ \\
        & $B$ & $\bot$, $\{P\}$, $\{Q\}$, $A$, $B$, $\{GO_1\}$, $\{GO_2\}$, $\top$ \\
        & $\{GO_1\}$ & $\bot$, $\{P\}$, $\{Q\}$, $A$, $B$, $\{GO_1\}$, $\{GO_2\}$, $\top$ \\
        & $\{GO_2\}$ & $\bot$, $\{P\}$, $\{Q\}$, $A$, $B$, $\{GO_1\}$, $\{GO_2\}$, $\top$ \\
        & $\top$ & $\bot$, $\{P\}$, $\{Q\}$, $A$, $B$, $\{GO_1\}$, $\{GO_2\}$, $\top$ \\
    \hline
    \multirow{7}{*}{$\{P\}$} & $\{P\}$ & $\{P\}$, $B$, $\top$ \\
        & $\{Q\}$ & $\bot$, $\{P\}$, $\{Q\}$, $A$, $B$, $\{GO_1\}$, $\{GO_2\}$, $\top$ \\
        & $A$ & $\bot$, $\{P\}$, $\{Q\}$, $A$, $B$, $\{GO_1\}$, $\{GO_2\}$, $\top$ \\
        & $B$ & $\{P\}$, $B$, $\top$ \\
        & $\{GO_1\}$ & $\{P\}$, $\{GO_1\}$, $B$, $\top$ \\
        & $\{GO_2\}$ & $\{P\}$, $\{GO_2\}$, $B$, $\top$ \\
        & $\top$ & $\{P\}$, $B$, $\top$ \\
    \hline
    \multirow{6}{*}{$\{Q\}$} & $\{Q\}$ & $\{Q\}$, $A$, $\top$ \\
        & $A$ & $\{Q\}$, $A$, $\top$ \\
        & $B$ & $\bot$, $\{P\}$, $\{Q\}$, $A$, $B$, $\{GO_1\}$, $\{GO_2\}$, $\top$ \\
        & $\{GO_1\}$ & $\{Q\}$, $\{GO_1\}$, $A$, $\top$ \\
        & $\{GO_2\}$ & $\{Q\}$, $\{GO_2\}$, $A$, $\top$ \\
        & $\top$ & $\{Q\}$, $A$, $\top$ \\
    \hline
    \multirow{5}{*}{$A$} & $A$ & $A$, $\top$ \\
        & $B$ & $\bot$, $\{P\}$, $\{Q\}$, $A$, $B$, $\{GO_1\}$, $\{GO_2\}$, $\top$ \\
        & $\{GO_1\}$ & $A$, $\{GO_1\}$, $\top$ \\
        & $\{GO_2\}$ & $A$, $\{GO_2\}$, $\top$ \\
        & $\top$ & $A$, $\top$ \\
    \hline
    \multirow{4}{*}{$B$} & $B$ & $B$, $\top$ \\
        & $\{GO_1\}$ & $B$, $\{GO_1\}$, $\top$ \\
        & $\{GO_2\}$ & $B$, $\{GO_2\}$, $\top$ \\
        & $\top$ & $B$, $\top$ \\
    \hline
    \multirow{3}{*}{$\{GO_1\}$} & $\{GO_1\}$ & $\{GO_1\}$, $\top$ \\
        & $\{GO_2\}$ & $\bot$, $\{P\}$, $\{Q\}$, $A$, $B$, $\{GO_1\}$, $\{GO_2\}$, $\top$ \\
        & $\top$ & $\{GO_1\}$, $\top$ \\
    \hline
    \multirow{2}{*}{$\{GO_2\}$} & $\{GO_2\}$ & $\{GO_2\}$, $\top$ \\
        & $\top$ & $\{GO_2\}$, $\top$ \\
    \hline
    $\top$ & $\top$ & $\top$ \\
  \end{tabular}
\end{adjustbox}
\end{center}

\section{Deductive Closure Computation Soundness}

Let us show that each inference rule provides truth statements:

\begin{enumerate}
    \item $\inference {A \sqcap B \sqsubseteq E \quad A' \sqsubseteq A \quad B' \sqsubseteq B \quad E \sqsubseteq E'}{A' \sqcap B' \sqsubseteq E'}$ \\
    Let $\mathcal{I}$ be an arbitrary interpretation. Since $\mathcal{I} \models A \sqcap B \sqsubseteq E$ then $A^{\mathcal{I}} \cap B^{\mathcal{I}} \subseteq E^{\mathcal{I}}$. Also, $A'^{\mathcal{I}} \subseteq A^{\mathcal{I}}$, $B'^{\mathcal{I}} \subseteq B^{\mathcal{I}}$ and $E^{\mathcal{I}} \subseteq E'^{\mathcal{I}}$. From this we derive $A'^{\mathcal{I}} \cap B'^{\mathcal{I}} \subseteq E'^{\mathcal{I}}$, i.e., $\mathcal{I} \models A' \sqcap B' \sqsubseteq E'$.
    \vspace{0.5cm}
    \item $\inference {A \sqsubseteq \exists r.B \quad A' \sqsubseteq A \quad B \sqsubseteq B' \quad r \sqsubseteq r'}{A' \sqsubseteq \exists r'.B'}$ \\
    Let $\mathcal{I}$ be an arbitrary interpretation. Since $\mathcal{I} \models A \sqsubseteq \exists r.B$ then $A^{\mathcal{I}} \subseteq \{a \in \Delta^{\mathcal{I}} | \exists b \in \Delta^{\mathcal{I}}: (a, b) \in r^{\mathcal{I}} \land b \in B^{\mathcal{I}}\}$. Additionally, $A'^{\mathcal{I}} \subseteq A^{\mathcal{I}}$, $B^{\mathcal{I}} \subseteq B'^{\mathcal{I}}$ and for arbitrary $a, b \in \Delta^{\mathcal{I}}$ $(a, b) \in r^{\mathcal{I}} \Rightarrow (a, b) \in r'^{\mathcal{I}}$. Then $A'^{\mathcal{I}} \subseteq \{a \in \Delta^{\mathcal{I}} | \exists b \in \Delta^{\mathcal{I}}: (a, b) \in r'^{\mathcal{I}} \land b \in B'^{\mathcal{I}}\}$, i.e., $\mathcal{I} \models A' \sqsubseteq \exists r'.B'$. 
    \vspace{0.5cm}
    \item $\inference {A \sqsubseteq \exists r.B \quad B \sqsubseteq \exists r'.E \quad r \circ r' \sqsubseteq s}{A \sqsubseteq \exists s.E}$ \\
    Let $\mathcal{I}$ be an arbitrary interpretation. Since $\mathcal{I} \models A \sqsubseteq \exists r.B$ and $\mathcal{I} \models B \sqsubseteq \exists r'.E$ then $A^{\mathcal{I}} \subseteq \{a \in \Delta^{\mathcal{I}} | \exists b \in \Delta^{\mathcal{I}}: (a, b) \in r^{\mathcal{I}} \land b \in B^{\mathcal{I}}\}$ and $B^{\mathcal{I}} \subseteq \{b \in \Delta^{\mathcal{I}} | \exists c \in \Delta^{\mathcal{I}}: (b, c) \in r'^{\mathcal{I}} \land c \in E^{\mathcal{I}}\}$. For arbitrary $a, b, c \in \Delta^{\mathcal{I}}$ $(a, b) \in r^{\mathcal{I}} \land (b, c) \in r'^{\mathcal{I}} \Rightarrow (a, c) \in s^{\mathcal{I}}$. Then $A^{\mathcal{I}} \subseteq \{a \in \Delta^{\mathcal{I}} | \exists c \in \Delta^{\mathcal{I}}: (a, c) \in s^{\mathcal{I}} \land c \in E^{\mathcal{I}}\}$, i.e., $\mathcal{I} \models A \sqsubseteq \exists s.E$. 
    \vspace{0.5cm} 
    \item $\inference {\exists r.A \sqsubseteq B \quad A' \sqsubseteq A \quad B \sqsubseteq B' \quad r' \sqsubseteq r}{\exists r'.A' \sqsubseteq B'}$ \\
    Let $\mathcal{I}$ be an arbitrary interpretation. Since $\mathcal{I} \models \exists r.A \sqsubseteq B$ then $\{a \in \Delta^{\mathcal{I}} | \exists b \in \Delta^{\mathcal{I}}: (a, b) \in r^{\mathcal{I}} \land b \in A^{\mathcal{I}}\} \subseteq D^{\mathcal{I}}$. Also, $A'^{\mathcal{I}} \subseteq A^{\mathcal{I}}$, $B^{\mathcal{I}} \subseteq B'^{\mathcal{I}}$ and for arbitrary $a, b \in \Delta^{\mathcal{I}}$ $(a, b) \in r'^{\mathcal{I}} \Rightarrow (a, b) \in r^{\mathcal{I}}$. From this follows $\{a \in \Delta^{\mathcal{I}} | \exists b \in \Delta^{\mathcal{I}}: (a, b) \in r'^{\mathcal{I}} \land b \in A'^{\mathcal{I}}\} \subseteq B'^{\mathcal{I}}$, i.e., $\mathcal{I} \models \exists r'.A' \sqsubseteq B'$. 
    \vspace{0.5cm}
    \item $\inference {A \sqcap B \sqsubseteq \bot \quad A' \sqsubseteq A \quad B' \sqsubseteq B}{A' \sqcap B' \sqsubseteq \bot}$ \\
    Let $\mathcal{I}$ be an arbitrary interpretation. Since $\mathcal{I} \models A \sqcap B \sqsubseteq \bot$ then $A^{\mathcal{I}} \cap B^{\mathcal{I}} = \emptyset$. Additionally, Also, $A'^{\mathcal{I}} \subseteq A^{\mathcal{I}}$, $B'^{\mathcal{I}} \subseteq B^{\mathcal{I}}$, from where we can derive $A'^{\mathcal{I}} \cap B'^{\mathcal{I}} = \emptyset$, i.e., $\mathcal{I} \models A' \sqcap B' \sqsubseteq \bot$.
    \vspace{0.5cm}
    \item $\inference {A \sqcap B \sqsubseteq \bot}{A \sqcap B \sqsubseteq E}$ \\
    Let $\mathcal{I}$ be an arbitrary interpretation. Since $\mathcal{I} \models A \sqcap B \sqsubseteq \bot$ then $A^{\mathcal{I}} \cap B^{\mathcal{I}} = \emptyset$. Since $\emptyset \subseteq E^{\mathcal{I}}$ for any concept $E$ then $A^{\mathcal{I}} \cap B^{\mathcal{I}} \subseteq E^{\mathcal{I}}$, i.e., $\mathcal{I} \models A \sqcap B \sqsubseteq E$. 
    \vspace{0.5cm}
    \item $\inference {\exists r.A \sqsubseteq \bot \quad A' \sqsubseteq A \quad r' \sqsubseteq r}{\exists r'.A' \sqsubseteq \bot}$ \\
    Let $\mathcal{I}$ be an arbitrary interpretation. Since $\mathcal{I} \models \exists r.A \sqsubseteq \bot$ then $\{a \in \Delta^{\mathcal{I}} | \exists b \in \Delta^{\mathcal{I}}: (a, b) \in r^{\mathcal{I}} \land b \in A^{\mathcal{I}}\} = \emptyset$. Additionally, $A'^{\mathcal{I}} \subseteq A^{\mathcal{I}}$ and for arbitrary $a, b \in \Delta^{\mathcal{I}}$ $(a, b) \in r'^{\mathcal{I}} \Rightarrow (a, b) \in r^{\mathcal{I}}$. From this we get that $\{a \in \Delta^{\mathcal{I}} | \exists b \in \Delta^{\mathcal{I}}: (a, b) \in r'^{\mathcal{I}} \land b \in A'^{\mathcal{I}}\} = \emptyset$, i.e., $\mathcal{I} \models \exists r'.A' \sqsubseteq \bot$.
    \vspace{0.5cm} 
    \item $\inference {}{A \sqcap \bot \sqsubseteq E}$ \\
    Let $\mathcal{I}$ be an arbitrary interpretation. For an arbitrary concept $A$ $(A \sqcap \bot)^{\mathcal{I}} = A^{\mathcal{I}} \cap \emptyset = \emptyset$ and for every concept $E$ we have $\emptyset \subseteq E^{\mathcal{I}}$. Hence $\mathcal{I} \models A \sqcap \bot \sqsubseteq E$. 
    \vspace{0.5cm} 
    \item $\inference {B \sqsubseteq \bot}{A \sqcap B \sqsubseteq E}$ \\
    Let $\mathcal{I}$ be an arbitrary interpretation. Since $\mathcal{I} \models B \sqsubseteq \bot$ then $B^{\mathcal{I}} = \emptyset$. On analogy with the previous case we get $\mathcal{I} \models A \sqcap B \sqsubseteq E$ for arbitrary concepts $A$ and $E$. 
    \vspace{0.5cm}
    \item $\inference {E \sqsubseteq E'}{A \sqcap E \sqsubseteq E'}$ \\
    Let $\mathcal{I}$ be an arbitrary interpretation. Since $\mathcal{I} \models E \sqsubseteq E'$ then $E^{\mathcal{I}} \subseteq E'^{\mathcal{I}}$. For every concept $A$ $(A \sqcap E)^{\mathcal{I}} = A^{\mathcal{I}} \cap E^{\mathcal{I}} \subseteq E^{\mathcal{I}} \subseteq E'^{\mathcal{I}}$, hence $\mathcal{I} \models A \sqcap E \sqsubseteq E'$. 
    \vspace{0.5cm}
    \item $\inference{A \sqcap B \sqsubseteq \bot}{A \sqcap B \sqsubseteq E}$ \\
    Let $\mathcal{I}$ be an arbitrary interpretation. Since $\mathcal{I} \models A \sqcap B \sqsubseteq \bot$ then $A^{\mathcal{I}} \cap B^{\mathcal{I}} = \emptyset$. Since $\emptyset \subseteq E^{\mathcal{I}}$ for every concept $E$ then $\mathcal{I} \models A \sqcap B \sqsubseteq E$. 
    \vspace{0.5cm}
    \item $\inference {A \sqsubseteq E \quad B \sqsubseteq E \quad A' \sqsubseteq A \quad B' \sqsubseteq B \quad E \sqsubseteq E'}{A' \sqcap B' \sqsubseteq E'}$ \\
    Let $\mathcal{I}$ be an arbitrary interpretation. Since $\mathcal{I} \models A \sqsubseteq E$ and $\mathcal{I} \models B \sqsubseteq E$ then $A^{\mathcal{I}} \subseteq E^{\mathcal{I}}$ and $B^{\mathcal{I}} \subseteq E^{\mathcal{I}}$. Note that $A^{\mathcal{I}} \cap B^{\mathcal{I}} \subseteq A^{\mathcal{I}} \subseteq E^{\mathcal{I}}$, from this we get $\mathcal{I} \models A \sqcap B \sqsubseteq E$. On analogy with case 1 we derive that $\mathcal{I} \models A' \sqcap B' \sqsubseteq E'$.
    \vspace{0.5cm} 
    \item $\inference{A \sqsubseteq A'}{A \sqcap \top \sqsubseteq A'}$ \\
    Let $\mathcal{I}$ be an arbitrary interpretation. Since $\mathcal{I} \models A \sqsubseteq A'$ then $A^{\mathcal{I}} \subseteq A'^{\mathcal{I}}$. By definition of $\top$, $(A \sqcap \top)^{\mathcal{I}} = A^{\mathcal{I}} \cap T^{\mathcal{I}} = A^{\mathcal{I}}$, hence $\mathcal{I} \models A \sqcap \top \sqsubseteq A'$. 
    \vspace{0.5cm}
    \item $\inference {}{\bot \sqsubseteq \exists r.B}$ \\
    Follows immediately from the fact that $\emptyset$ is a subset of any concept interpretation.
    \vspace{0.5cm}
    \item $\inference {A \sqsubseteq \bot}{A \sqsubseteq \exists r.B}$ \\
    Let $\mathcal{I}$ be an arbitrary interpretation. Since $\mathcal{I} \models A \sqsubseteq \bot$ then $A^{\mathcal{I}} = \emptyset$. On analogy with case 14 we get $\mathcal{I} \models A \sqsubseteq \exists r.B$. 
    \vspace{0.5cm}
    \item $\inference {}{\exists r.A \sqsubseteq \top}$ \\
    Follows immediately from the fact that $\top^{\mathcal{I}} = \Delta^{\mathcal{I}}$. 
\end{enumerate}

\end{document}